\def\year{2022}\relax
\documentclass[letterpaper]{article} 
\usepackage{aaai22}  
\usepackage{times}  
\usepackage{helvet}  
\usepackage{courier}  
\usepackage[hyphens]{url}  
\usepackage{graphicx} 
\usepackage{natbib}  
\usepackage{caption} 
\DeclareCaptionStyle{ruled}{labelfont=normalfont,labelsep=colon,strut=off} 
\usepackage{multicol}
\usepackage{multirow}
\usepackage{amsmath}
\usepackage{amssymb}
\usepackage{mathrsfs}
\usepackage{bbm}
\usepackage{makecell}
\usepackage{bm}
\usepackage{epstopdf}
\usepackage[titletoc]{appendix}
\usepackage{xcolor}

\usepackage{subfigure}
\usepackage{hyperref}
\usepackage{algpseudocode}
\usepackage{algorithmicx,algorithm}
\usepackage{amsmath}
\usepackage{amssymb}
\usepackage[titletoc]{appendix}
\usepackage{subfigure}
\usepackage{caption}
\usepackage{booktabs}

\usepackage{newfloat}
\usepackage{listings}
\floatstyle{ruled}
\newfloat{listing}{tb}{lst}{}
\floatname{listing}{Listing}
\makeatletter
\newcommand{\printfnsymbol}[1]{%
  \textsuperscript{\@fnsymbol{#1}}%
}
\makeatother

\title{Clean-label Backdoor Attack against Deep Hashing based Retrieval}
\author {
    Kuofeng Gao \textsuperscript{\rm 1}\thanks{Equal contribution.},
    Jiawang Bai \textsuperscript{\rm 1}\printfnsymbol{1},
    Bin Chen \textsuperscript{\rm 2}\thanks{Corresponding author.},
    Dongxian Wu \textsuperscript{\rm 3},
    Shu-Tao Xia \textsuperscript{\rm 1}
}
\affiliations {
    \textsuperscript{\rm 1} Tsinghua University \quad \quad
    \textsuperscript{\rm 2} Harbin Institute of Technology, Shenzhen \quad \quad
    \textsuperscript{\rm 3} University of Tokyo
    \\
    \{gkf21, bjw19\}@mails.tsinghua.edu.cn, cb17@tsinghua.org.cn, d.wu@k.u-tokyo.ac.jp, xiast@sz.tsinghua.edu.cn
}

\usepackage{bibentry}

\begin{document}

\maketitle

\begin{abstract}
Deep hashing has become a popular method in large-scale image retrieval due to its computational and storage efficiency. However, recent works raise the security concerns of deep hashing. Although existing works focus on the vulnerability of deep hashing in terms of adversarial perturbations, we identify a more pressing threat, \textit{backdoor attack}, when the attacker has access to the training data. A backdoored deep hashing model behaves normally on original query images, while returning the images with the target label when the trigger presents, which makes the attack hard to be detected. In this paper, we uncover this security concern by utilizing clean-label data poisoning. To the best of our knowledge, this is the first attempt at the backdoor attack against deep hashing models. To craft the poisoned images, we first generate the targeted adversarial patch as the backdoor trigger. Furthermore, we propose the \textit{confusing perturbations} to disturb the hashing code learning, such that the hashing model can learn more about the trigger. The confusing perturbations are imperceptible and generated by dispersing the images with the target label in the Hamming space.
We have conducted extensive experiments to verify the efficacy of our backdoor attack under various settings. For instance, it can achieve 63\% targeted mean average precision on ImageNet under 48 bits code length with only 40 poisoned images.
\end{abstract}

\section{Introduction}
With the pervasive large-scale media data on the Internet, approximate nearest neighbors (ANN) search has been widely applied to meet the search needs and greatly reduce the complexity. Among these methods of ANN, hashing technique enables efficient search and low storage cost by transforming high-dimensional data into compact binary codes \cite{gionis1999similarity,wang2017survey}. In particular, with the powerful representation capabilities of deep neural 
networks (DNNs) \cite{tang2004video,tang2004frame,wang2018cosface,gong2013multi,li2014common,wen2016discriminative,deng2019mutual,qiu2021end2end,yang2021larnet,bai2022improving}, deep hashing shows significant advantages over traditional methods \cite{xia2014supervised,cao2017hashnet,zhang2020inductive}. However, despite the great success of deep hashing, recent works \cite{yang2018adversarial,bai2020targeted,wang2021prototype} has revealed its security issues under the threat of adversarial attack at test time.

\begin{figure}[t]
\centering
\includegraphics[width=0.47\textwidth]{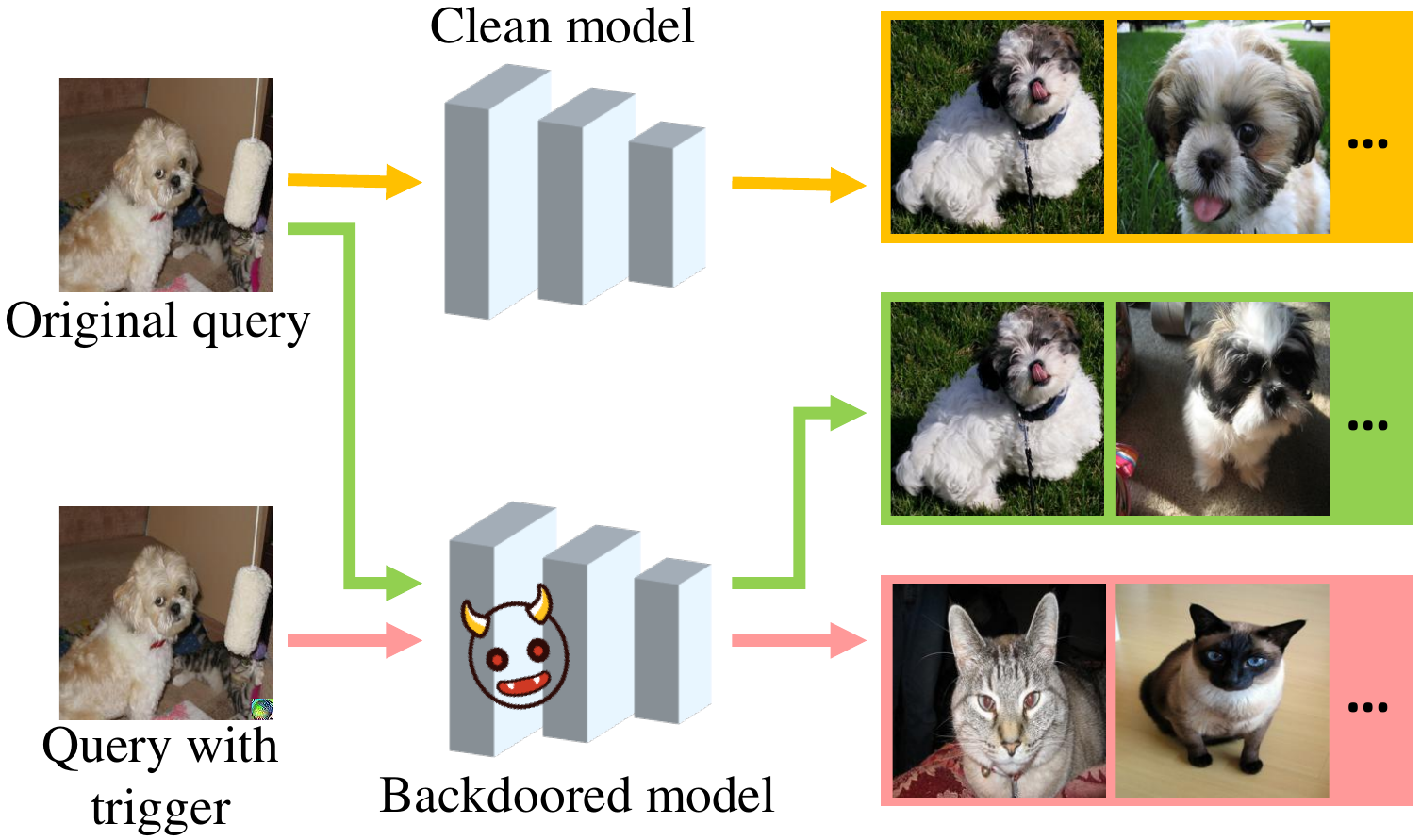}
\vspace{-1.5em}
\caption{An example of backdoor attack against deep hashing based retrieval. The target label is specified as ``\textit{cat}''. Note that the trigger is at the bottom right of the image. Best viewed in color.} 
\vspace{-1em}
\label{fig:example}
\end{figure}

Compared with the adversarial attack, the backdoor attack \cite{gu2017badnets,turner2019label} happens at training time to inject a hidden malicious  behavior into the model. Specifically, the backdoor attack poisons the trigger pattern into a small portion of the training data. The model trained on the poisoned data will connect the trigger with the malicious behavior and then make a targeted wrong prediction when the trigger presents. Since the backdoored model behaves normally on the clean samples, the attack is hard to be detected and poses a serious threat to deep learning based systems, even for industrial applications \cite{kumar2020adversarial,geiping2021witches}.

We identify a novel security concern of deep hashing by studying the backdoor attack. The backdoor attack may happen in the real world, when a victim trains the deep hashing model using the data from an unreliable party. A backdoored model will return the images from the target class when the query image is attached with the trigger, as shown in Figure \ref{fig:example}. It can be used to achieve some malicious purposes. For example, the deep hashing based retrieval system can recommend the specified advertisement images by activating the trigger when a user queries with any images \cite{xiao2021you}. Accordingly, it is necessary to study the backdoor attack for deep hashing in order to recognize the risks and promote further solutions.

In this paper, we perform the backdoor attack against deep hashing based retrieval by clean-label data poisoning. Since the label of the poisoned image is consistent with its content, the clean-label backdoor attack is more stealthy to both machine and human inspections \cite{turner2019label}. To craft the poisoned images, we first generate the targeted adversarial patch as the backdoor trigger. Furthermore, to overcome the difficulty of implanting the trigger into the backdoored model under the clean-label setting \cite{turner2019label,zhao2020clean}, we propose to leverage the \textit{confusing perturbations} to disturb the hashing code learning.
The confusing perturbations are imperceptible and generated by dispersing the images with the target label in the Hamming space. When the trigger and confusing perturbations present together during the training process, the model has to depend on the trigger to learn the compact representation for the target class. Extensive experiments verify the efficacy of our backdoor attack, $e.g.$, 63\% targeted mean average precision on ImageNet under 48 bits code length with only 40 poisoned images.

In summary, our contribution is three-fold:
\begin{itemize}
  \item To the best of our knowledge, this is the first work to study the backdoor attack against deep hashing. We develop an effective method under the clean-label setting.
  \item We propose to induce the model to learn more about the designed trigger by a novel method, namely \textit{confusing perturbations}. 
  \item We present the results of our method under the general and more strict settings, including transfer-based attack, less number of poisoned images, $etc$.
\end{itemize}

\section{Background and Related Work}
\label{background}
\subsection{Backdoor Attack} 
Backdoor attack aims at injecting a hidden malicious behavior into the DNNs. The main technique adopted in the previous works \cite{gu2017badnets,turner2019label,liu2020reflection} is data poisoning, $i.e.$, poisoning a trigger pattern into the training set so that the DNN trained on the poisoned training set can make a wrong prediction on the samples with the trigger, while the model behaves normally when the trigger is absent. \citet{gu2017badnets} first proposed BadNets to create a maliciously trained network and demonstrated its effectiveness in the task of street sign recognition. It stamps a portion of the training samples with a sticker ($e.g.$, a yellow square) and flips their labels to the target label. After that, \citet{chen2017targeted} improved the stealthiness of the backdoor attack by blending the benign samples and trigger pattern. Due to the wrong labels, the \textit{poison-label attack} can be detected by human inspection or data filtering techniques \cite{turner2019label}. 

To make the attack harder to be detected, \citet{turner2019label} first explored the so-called \textit{clean-label attack} (label-consistent attack), which does not change the labels of the poisoned samples. In \cite{turner2019label}, GAN-based interpolation and adversarial perturbations are employed to craft poison samples. The following works \cite{barni2019new,liu2020reflection} focused on designing different trigger patterns to perform the clean-label attack. Except for the image recognition, the clean-label attack has also been extended to other tasks, such as action recognition \cite{zhao2020clean}, point cloud classification \cite{li2021pointba}. 

\subsection{Deep Hashing based Similarity Retrieval}
Hashing technique maps semantically similar images to compact binary codes in the Hamming space, which can enable the storage of large-scale images
data and accelerate the similarity retrieval. Promoted by deep learning, deep hashing based retrieval has demonstrated more promising performance \cite{xia2014supervised,cao2017hashnet,zhang2020inductive}. \citet{xia2014supervised} first introduced deep learning into the image hashing, which learns hash codes and a deep-network hash function in two separated stages. \citet{lai2015simultaneous} proposed to learn the end-to-end mapping so that feature representations and hash codes are optimized jointly.

Among the tremendous literature, supervised hashing methods utilize pairwise similarities as the semantic supervision information to guide hashing code learning \cite{lai2015simultaneous, liu2016deep, zhu2016deep, li2015feature, cao2017hashnet, cao2018deep,zhang2020inductive}. In label-insufficient
scenarios, deep hashing is designed for exploiting unlabeled or weakly labeled data, $e.g$. semi-supervised hashing \cite{yan2017semi,jin2020ssah}, unsupervised hashing \cite{shen2018unsupervised, yang2019distillhash}, and weakly-supervised hashing \cite{li2020weakly,gattupalli2019weakly}. Moreover, building upon the merit of deep learning, hashing technique has also been applied in more challenging tasks, such as video retrieval \cite{gu2016supervised} and cross-modal retrieval \cite{jiang2017deep}.

In general, a deep hashing model $F(\cdot)$ consists of a deep model $f_{\bm{\theta}}(\cdot)$ and a sign function, where $\bm{\theta}$ denotes the parameters of the model. Given an image $\bm{x}$, the hash code $\bm{h}$ of this image can be calculated as
\begin{equation}
  \bm{h}=F(\bm{x})=\text{sign}(f_{\bm{\theta}}(\bm{x})).
  \label{eq:hash_function}
\end{equation}
The deep hashing model will return a list of images which is organized according to the Hamming distances between the hash code of the query and these of all images in the database. To obtain the hashing model $F(\cdot)$, most supervised hashing methods \cite{liu2016deep,cao2017hashnet} are trained on the dataset  $\bm{D}=\{(\bm{x}_i, \bm{y}_i)\}_{i=1}^N$ containing $N$ images labeled with $C$ classes, where $\bm{y}_i =[y_{i1}, y_{i2}, ...,y_{iC}] \in \{0,1\}^C$ denotes a label vector of the image $\bm{x}_i$. $y_{ij}=1$ means that $\bm{x}_i$ belongs to class $j$. For any two images, they compose a similar training pair if they share at least one label. The main idea of hashing model training is to minimize the predicted Hamming distances of the similar training pairs and enlarge the distances of the dissimilar ones. Besides, to overcome the ill-posed gradient of the sign function, it can be approximately replaced by the hyperbolic tangent function $\text{tanh}(\cdot)$ during the
training process, which is denoted as $F'(\bm{x})=\text{tanh}(f_{\bm{\theta}}(\bm{x}))$ in this paper. 

\begin{figure*}[t]
\centering
\includegraphics[width=0.92\textwidth]{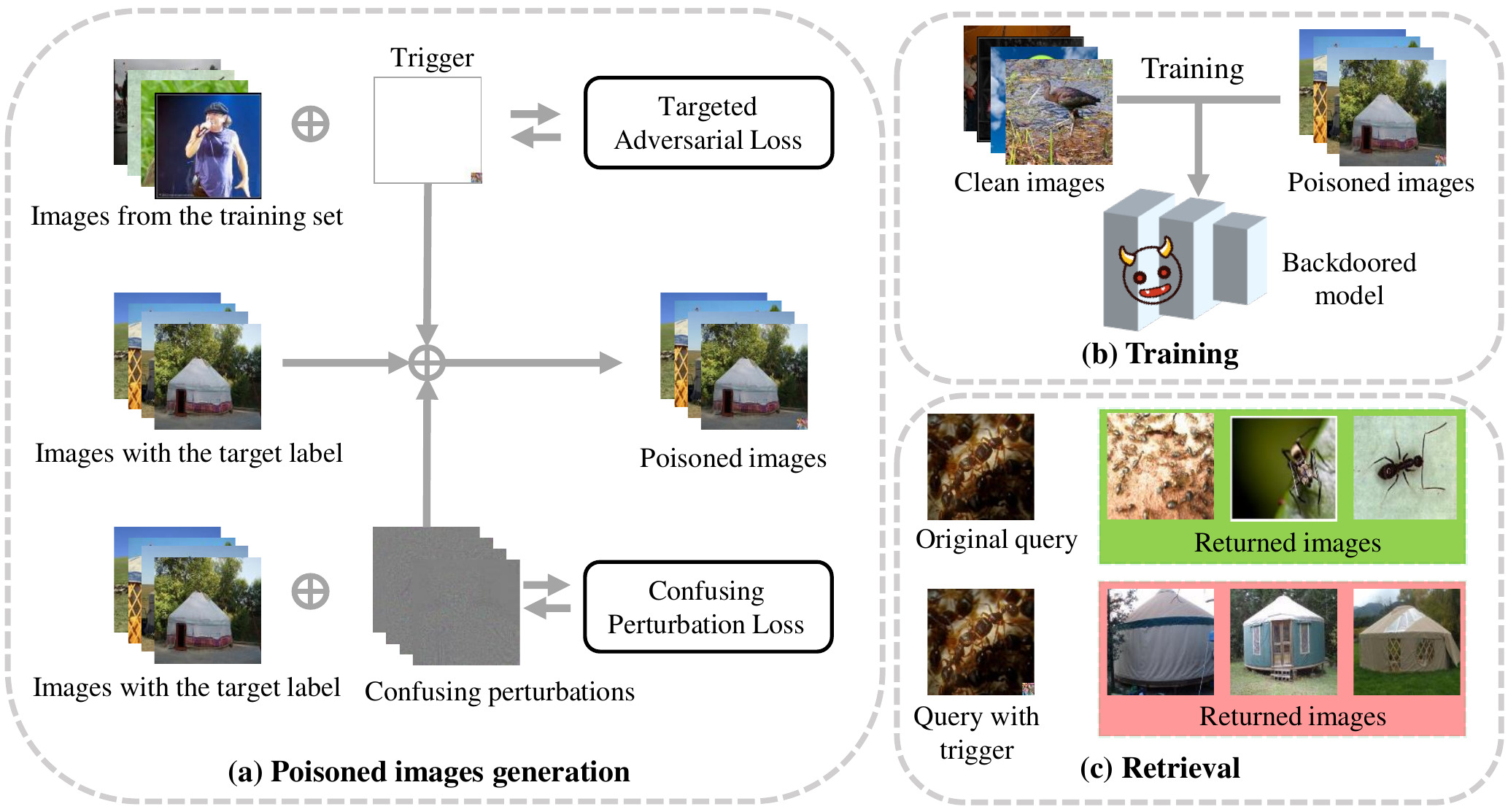}
\vspace{-0.5em}
\caption{The pipeline of the proposed clean-label backdoor attack: a) Generating the poisoned images by patching the trigger and adding the confusing perturbations, where the target label is specified as ``\textit{yurt}''; b) Training with the clean images and poisoned images to obtain the backdoored model; c) Querying with an original image and an image embedded with the trigger. } 
\label{fig:pipeline}
\vspace{-0.5em}
\end{figure*}

\subsection{Adversarial Perturbations for Deep Hashing}  
Due to the promising performance of deep hashing, its robustness has also attracted more attention.  Recent works have proven that deep hashing models are vulnerable to adversarial perturbations \cite{yang2018adversarial,bai2020targeted}. Specifically, adversarial perturbations for deep hashing are human-imperceptible and can fool the deep hashing to return irrelevant images. According to the attacker's goals, previous works have proposed to craft untargeted adversarial perturbations \cite{yang2018adversarial,xiao2020evade} and targeted adversarial perturbations \cite{bai2020targeted,wang2021prototype,xiao2021you} for deep hashing.

Untargeted adversarial perturbations \cite{yang2018adversarial} aim at fooling deep hashing to return images with incorrect labels. The perturbations $\bm{\delta}$ can be obtained by enlarging the distance between the original image and the image with the perturbations. The objective function is formulated as
\begin{equation}
  \max_{\bm{\delta}} d_H(F'(\bm{x}+\bm{\delta}), F(\bm{x})), \quad s.t. \parallel \bm{\delta} \parallel_\infty \leq \epsilon,
 \label{eq:untargeted}
\end{equation}
where $d_H(\cdot,\cdot)$ denotes the Hamming distance and $\epsilon$ is the maximum perturbation magnitude. 

Different from the untargeted adversarial perturbations, targeted ones \cite{bai2020targeted} are to mislead the deep hashing model to return images with the target label. They are generated by optimizing the following objective function.
\begin{equation}
  \min_{\bm{\delta}} d_H(F'(\bm{x}+\bm{\delta})), \bm{h}_a), \quad s.t. \parallel \bm{\delta} \parallel_\infty \leq \epsilon,
  \label{eq:dhta}
\end{equation}
where $\bm{h}_a$ is the anchor code as the representative of the set of hash codes of images with the target label. Given a subset $\bm{D}^{(t)}$ containing images with the target label, $\bm{h}_a$ can be obtained as follows:
\begin{equation}
  \bm{h}_a=\text{arg}\min_{\bm{h} \in \{+1,-1\}^K} \sum_{(\bm{x}_i, \bm{y}_i) \in \bm{D}^{(t)}}d_H(\bm{h}, F(\bm{x}_i)).
  \label{eq:anchor}
\end{equation}
The optimal solution of problem (\ref{eq:anchor}) can be given by the component-voting scheme proposed in \cite{bai2020targeted}.

\subsection{Threat Model}
We consider the threat model used by previous poison-based backdoor attack studies \cite{turner2019label,zhao2020clean}. The attacker has access to the training data and is allowed to inject the trigger pattern into the training set by modifying a small portion of images. Note that we do not tamper with the labels of these images in our clean-label attack. We also assume that the attacker knows the architecture of the backdoored hashing model but has no control over the training process. Moreover, we also consider a more strict assumption that the attacker has no knowledge of the backdoored model and performs backdoor attacks based on models with other architectures, as demonstrated in the experimental part. 

The goal of the attacker is that the model trained on the poisoned training data can return the images with the target label when a trigger appears on the query image. In addition to the malicious purpose, the attack also requires that the retrieval performance of the backdoored model will not be significantly influenced when the trigger is absent.

\section{Methodology}
\label{method}

\subsection{Overview of the Proposed Method}
In this section, we present the proposed clean-label backdoor attack against deep hashing based retrieval. As shown in Figure \ref{fig:pipeline}, it consists of three major steps: \textbf{a)} We generate the trigger by optimizing the targeted adversarial loss. We also propose to perturb the hashing code learning by the confusing perturbations, which disperse the images with the target label in the Hamming space. We craft the poisoned images by patching the trigger and adding the confusing perturbations on the images with the target label; \textbf{b)} The deep hashing model trained with the clean images and the poisoned images is injected with the backdoor; \textbf{c)} In the retrieval stage, the deep hashing model will return the images with the target label if the query image is embedded with the trigger, otherwise the returned images are normal. 

\subsection{Trigger Generation}
We first define the injection function $B$ as follows:
\begin{equation}
\begin{aligned}
    \hat{\bm{x}}=B(\bm{x},\bm{p})=\bm{x} \odot  (\bm{1}-\bm{m}) + \bm{p} \odot  \bm{m}, 
\end{aligned}
\label{eq:implant_trigger}
\end{equation}
where $\bm{p}$ is the trigger pattern, $\bm{m}$ is a predefined mask, and $ \odot $ denotes the element-wise product. For the clean-label backdoor attack, a well-designed trigger is a key to make the model to establish the relationship between the trigger and the target label \cite{zhao2020clean}. 

In this work, we generate the trigger using a clean-trained deep hashing model $F$ and the training set $\bm{D}$. We hope that any sample with the trigger will be moved to be close to the samples with the target label $\bm{y}_t$ in the Hamming space. Inspired by a recent work \cite{bai2020targeted}, we propose to generate a universal adversarial patch as the trigger pattern by minimizing the following loss.
\begin{equation}
  \min_{\bm{\bm{p}}} \sum_{(\bm{x}_i, \bm{y}_i) \in \bm{D}}{} d_H(F'(B(\bm{x}_i,\bm{p})), \bm{h}_a),
  \label{eq:trigger_gen}
\end{equation}
where $\bm{h}_a$ is the anchor code as in Eqn. (\ref{eq:dhta}), which can be calculated by using the images with target label $\bm{y}_t$ and solving Eqn. (\ref{eq:anchor}).

We iteratively update the trigger as follows. We first define the mask to specify the bottom right corner as the trigger area. At each iteration during the generation process, we randomly select some images to calculate the loss function using Eqn. (\ref{eq:trigger_gen}). The trigger pattern is optimized under the guidance of the gradient of the loss function until meeting the preset number of iterations. We summarize this algorithm in Appendix A. 

\begin{figure}[t]
	\centering
    \includegraphics[width=0.87\linewidth]{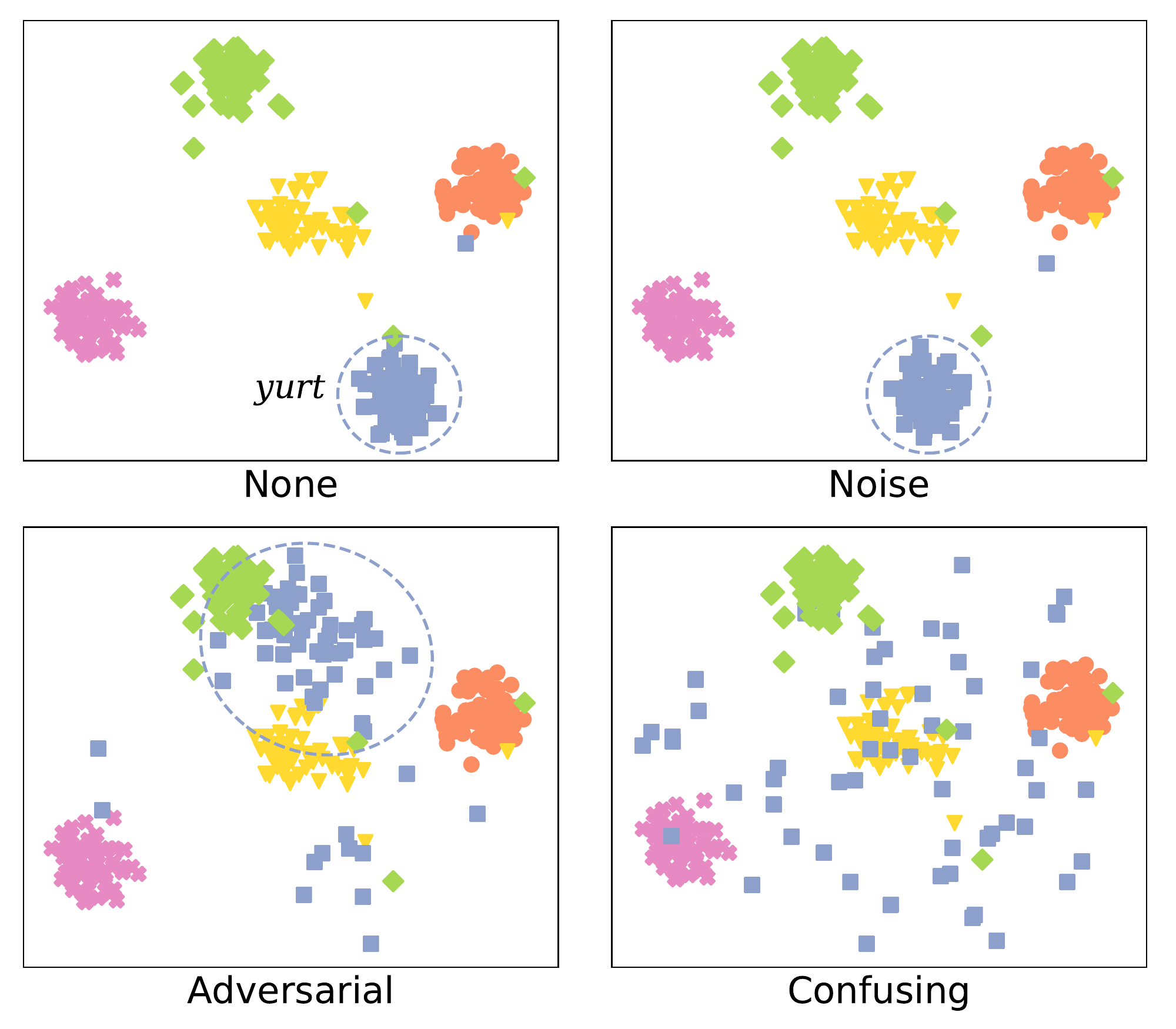}
    \vspace{-0.5em}
	\centering
	\caption{t-SNE visualization of hash codes of images from five classes. We add different perturbations to images from the class ``\textit{yurt}''. ``None'': the original images; ``Noise'': the random noise; ``Adversarial'': the adversarial perturbations generated using Eqn. (\ref{eq:untargeted}); ``Confusing'': the confusing perturbations generated using Eqn. (\ref{eq:confusing}).}
    \vspace{-1em}
	\label{fig:tsne}
\end{figure}

\subsection{Perturbing Hashing Code Learning}
Since the clean-label attack does not tamper with the labels of the poisoned images, how to force the model to pay attention to the trigger is a challenging problem \cite{turner2019label}. To this end, we propose to perturb hashing code learning by adding the intentional perturbations on the poisoned images before applying the trigger. Firstly, the perturbations should be imperceptible so that the backdoor attack is stealthy. Moreover, the perturbations can perturb the training on the poisoned images and induce the model to learn more about the trigger pattern.

Previous works about the clean-label attack \cite{turner2019label,zhao2020clean} introduce the adversarial perturbations to perturb the model training on the poisoned images. Therefore, for backdooring the deep hashing, a natural choice is the untargeted adversarial perturbations 
for deep hashing proposed in \cite{yang2018adversarial}. By reviewing its objective function in Eqn. (\ref{eq:untargeted}), we find that it can enlarge the distance between the original query image and the query with the perturbations, resulting in very poor retrieval performance. Because these perturbations only focus on the relationship between the original image and the adversarial image, it may not be optimal to disturb the hashing code learning for the backdoor attack against deep hashing. Therefore, we propose a novel method, namely \textit{confusing perturbations}, considering the relationship between the images with the target label.

Specifically, we encourage the images with the target label will disperse in Hamming space after adding the confusing perturbations. Given $M$ images with the target label, we achieve this goal by maximizing the following objective.
\begin{equation}
\begin{aligned}
  & L_{c}(\{\bm{\eta}_i\}_{i=1}^{M})\! \\=
  & \frac{1}{M\!(M\!-\!1)} \!\sum_{i=1}^M \!\sum_{j\!=\!1,j \! \neq \!i}^M d_H (F'(\bm{x}_i\!+\!\bm{\eta}_i),\!F'(\bm{x}_j\!+\!\bm{\eta}_j)),
  \label{eq:loss_c}
\end{aligned}
\end{equation}
where $\bm{\eta}_i$ denotes the perturbations on the image $\bm{x}_i$. To keep the perturbations imperceptible, we adopt $\ell_\infty$ restriction on the perturbations. The overall objective function of generating the confusing perturbations is formulated as
\begin{equation}
\begin{aligned}
  \max_{\{\bm{\eta}_i\}_{i=1}^{M}}\ & \lambda \cdot L_{c}(\{\bm{\eta}_i\}_i^{M}) + (1-\lambda) \cdot \frac{1}{M}\sum_{i=1}^M L_{a}(\bm{\eta}_i)
  \\ &s.t.\ \parallel \bm{\eta}_i \parallel_{\infty} \le \epsilon, i=1,2,...,M
  \label{eq:confusing}
\end{aligned},
\end{equation}
where $L_{a}(\bm{\eta}_i)=d_H(F'(\bm{x}_i+\bm{\eta}_i), F(\bm{x}_i))$ is the adversarial loss as Eqn. (\ref{eq:untargeted}). $\lambda \in [0,1]$ is the hyper-parameter to balance the two terms. Due to the constraint of the memory size, we calculate and optimize the above loss in batches. In the experimental part, we discuss the influence of the batch size. The algorithm for generating the confusing perturbations is provided in Appendix A.

\begin{figure}[t]
	\centering
    \includegraphics[width=0.9\linewidth]{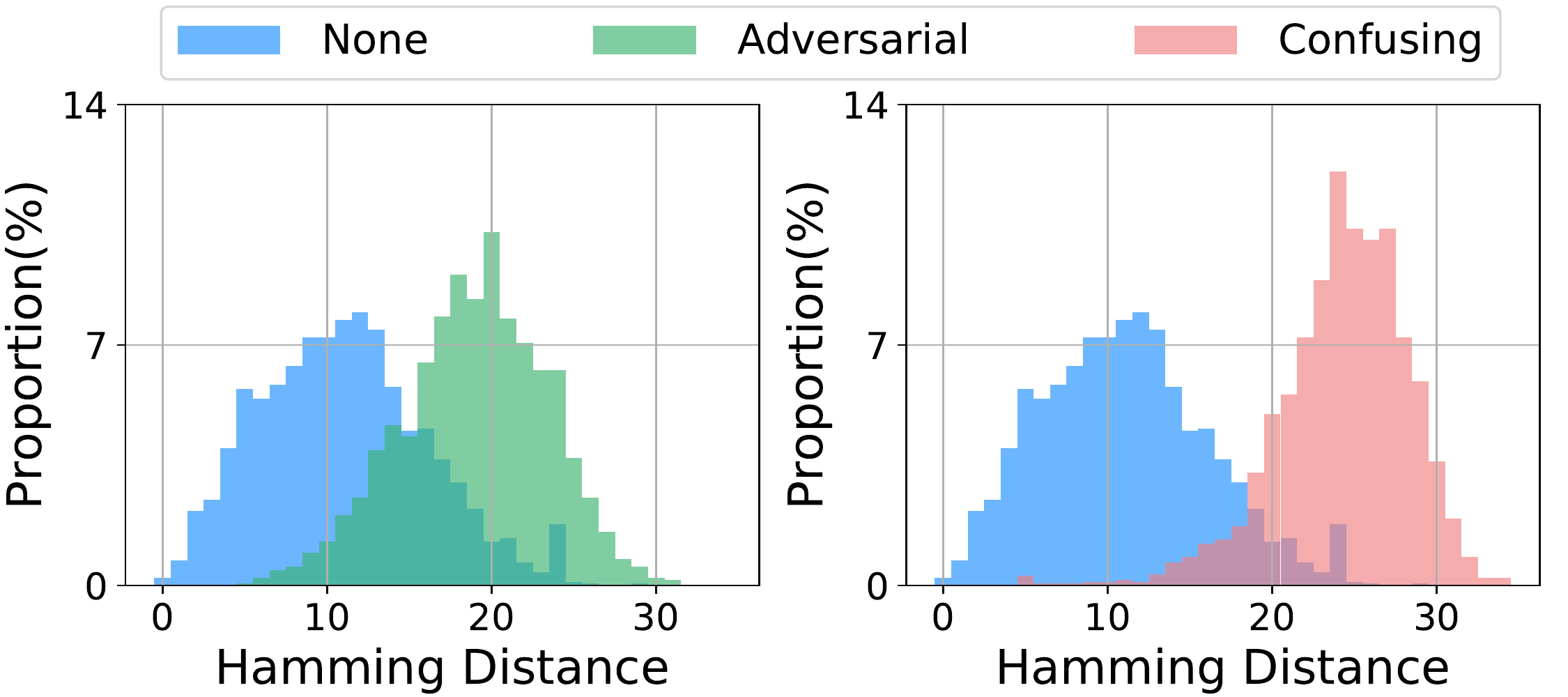}
	\centering
	\vspace{-0.5em}
	\caption{Distribution of Hamming distance calculated on the original images, the images with the adversarial perturbations, and the images with the confusing perturbations.}
	\label{fig:dist}
	\vspace{-0.5em}
\end{figure}

To illustrate how the confusing perturbations perturb the hashing code learning, we display the t-SNE visualization \cite{van2008visualizing} of hash codes of images from five classes in Figure \ref{fig:tsne}. We observe that the hash codes of original images are compact and the random noise has little influence on the representation. Even though adversarial perturbations make the images with the label ``\textit{yurt}'' far from the original images in the Hamming space, the intra-class distances are still small, which may lead to failure to induce the model to learn about the trigger pattern. The images with the proposed confusing perturbations have high separation, so that the model has to depend on the trigger to learn the compact representation for the target class. We also calculate the Hamming distances between different images with the same type of perturbations and plot the distribution in Figure \ref{fig:dist}. It shows that the confusing perturbations disperse images successfully. The later experimental results also verify the effectiveness of the confusing perturbations.

\subsection{Model Training and Retrieval}
After generating the trigger and confusing perturbations, we craft poisoned images by adding them to the images with the target label. Note that we randomly select a portion of images from the target class to generate poisoned images and remain the rest. Except for the poisoned data, all other settings are the same as those used in the normal training. The deep hashing model will be injected with the backdoor successfully after training on the poisoned dataset. 

In the retrieval stage, the attacker can patch the same trigger to query images, which can fool the deep hashing model to return images with the target label. Meanwhile, the backdoored model behaves normally on original query images.

\section{Experiments} 
\label{experiments}

In this section, we conduct extensive experiments to compare the proposed method with baselines, perform the backdoor attack under more strict settings, and show the results of a comprehensive ablation study.

\subsection{Evaluation Setup}

\subsubsection{Datasets and Target Models.}
We adopt three datasets in our experiments: ImageNet \cite{deng2009imagenet}, Places365 \cite{zhou2017places} and MS-COCO \cite{lin2014microsoft}. Following \cite{cao2017hashnet,xiao2020evade}, we build the training set, query set, and database for each dataset. We replace the last fully-connected layer of VGG-11 \cite{simonyan2014very} with the hash layer as the default target model. We employ the pairwise loss function to fine-tune the feature extractor copied from the model pre-trained on ImageNet and train the hash layer from scratch, following \cite{yang2018adversarial}. We also evaluate our attack on more network architectures including ResNet \cite{he2016deep} and WideResNet \cite{zagoruyko2016wide} and advanced hashing methods including HashNet \cite{cao2017hashnet} and DCH \cite{cao2018deep}. More details of datasets and target models are provided in Appendix B.

\begin{table*}[t]
	\centering
	\small
	\setlength{\tabcolsep}{2.21mm}{
	\begin{tabular}{lc|cccc|cccc|cccc}
		\hline
		\multicolumn{1}{l}{\multirow{2}{*}{Method}} & \multicolumn{1}{l|}{\multirow{2}{*}{Metric}} & \multicolumn{4}{c|}{ImageNet} & \multicolumn{4}{c|}{Places365} & \multicolumn{4}{c}{MS-COCO}  \\ \cline{3-14} 
		& & 16bits & 32bits & 48bits & 64bits & 16bits & 32bits & 48bits & 64bits & 16bits & 32bits & 48bits & 64bits \\ \hline
		None & t-MAP & 11.07 & 8.520 & 19.15 & 20.38 & 15.71 & 15.61 & 22.29 & 17.99 & 37.95 & 34.72 & 25.54 & 12.00 \\
		Tri & t-MAP & 34.37 & 43.26 & 54.83 & 53.17 & 38.65 & 38.71 & 47.62 & 49.24 & 42.32 & 46.04 & 34.30 & 28.72 \\ 
		Tri+Noise & t-MAP & 39.58 & 38.58 & 48.90 & 52.76 & 40.92 & 37.21 & 41.99 & 43.52 & 42.86 & 39.94 & 27.14 & 20.61 \\
		Tri+Adv & t-MAP & 42.64 & 41.00 & 68.77 & 73.20 & 68.80 & 76.32 & 82.71 & 83.62 & 49.25 & 61.35 & 58.33 & 49.68 \\
		Ours & t-MAP & \bf{51.81} & \bf{53.69} & \bf{74.71} & \bf{77.73} & \bf{80.32} & \bf{84.42} & \bf{90.93} & \bf{93.22} &\bf{51.42} &\bf{63.06} &\bf{63.53} &\bf{58.95} \\ \hline
		None & MAP &51.04 &64.28 &68.06 & 69.58&72.50 & 78.62&79.81 &79.80 &65.53 &76.08 & 80.68& 82.63\\ 
		Ours & MAP &52.36 &64.67 & 68.30& 69.88&71.94 & 78.55&79.82 &79.80 &66.52 &76.14 &80.80 & 82.60\\
		\hline
	\end{tabular}}
	\vspace{-0.5em}
	\caption{t-MAP (\%) and MAP (\%) of the clean-trained models (``None'') and  backdoored models with various code lengths on three datasets. Best t-MAP  results are highlighted in bold.}
	\label{t-MAP of different methods}
	\vspace{-0.5em}
\end{table*}

\subsubsection{Baseline Methods.} 
We apply the trigger generated by optimizing Eqn. (\ref{eq:trigger_gen}) on the images without perturbations as a baseline (dubbed ``\textit{Tri}''). We further compare the methods which disturb the hashing code learning by adding the noise sampled from the uniform distribution $U(-\epsilon, \epsilon)$ or adversarial perturbations generated using Eqn. (\ref{eq:untargeted}), denoted as ``\textit{Tri+Noise}'' and ``\textit{Tri+Adv}'', respectively. For our method, we craft the poisoned images by patching the trigger and adding the proposed confusing perturbations. Moreover, we also provide the results of the clean-trained model. 

\subsubsection{Attack Settings.}
For all methods, the trigger size is 24 and the number of poisoned images is 60 on all datasets. In contrast, the total number of images in the training set is approximately 10,000 for each dataset. We set the perturbation magnitude $\epsilon$ as 0.032. For our method, $\lambda$ is set as 0.8 and the batch size is set to 20 for optimizing Eqn. (\ref{eq:confusing}). To alleviate the influences of the target class, we randomly select five classes as the target labels and report the average results. Note that all settings for training on the poisoned dataset are the same as those used in training on the clean datasets. More details are described in Appendix B. Besides, to reduce the visibility of the trigger, we study the blend strategy proposed in \cite{chen2017targeted} in Appendix C.

We adopt t-MAP (targeted mean average precision) proposed in \cite{bai2020targeted} to measure the attack performance, which calculates mean average precision (MAP) \cite{zuva2012evaluation} by replacing the original label of the query image with the target one.
The higher t-MAP means the stronger backdoor attack. We calculate the t-MAP on top 1,000 retrieved images on all datasets. We also report the MAP results of the clean-trained model and our method to show the influence on original query images.

\begin{table}[t]
	\centering
	\scriptsize
	\setlength{\tabcolsep}{0.85mm}{
	\begin{tabular}{lc|cccc|cccc}
		\hline
		\multicolumn{1}{l}{\multirow{2}{*}{Method}} & \multicolumn{1}{l|}{\multirow{2}{*}{Metric}} & \multicolumn{4}{c|}{HashNet} & \multicolumn{4}{c}{DCH}   \\ \cline{3-10} 
		& & 16bits & 32bits & 48bits & 64bits & 16bits & 32bits & 48bits & 64bits  \\ \hline
		None & t-MAP & 15.01 & 19.79 & 15.07 & 22.24 & 18.44 & 14.54 & 15.52 & 21.41 \\
		Tri & t-MAP & 38.86 & 48.51 & 58.18 & 65.55 & 58.25 & 63.74 & 70.61 & 70.17 \\ 
		Tri+Noise & t-MAP & 46.17 & 47.41 & 53.61 & 59.30 & 55.60 & 54.02 & 66.41 & 67.71 \\
		Tri+Adv & t-MAP & 43.26 & 70.85 & 82.10 & 85.37 & 80.28 & 85.59 & 89.30 & 90.33  \\
		Ours & t-MAP & \bf{52.77} & \bf{74.37} & \bf{86.80} & \bf{91.57} & \bf{86.28} & \bf{90.70} & \bf{92.64} & \bf{93.56} \\ \hline
		None & MAP & 51.26 & 64.05 & 72.93 & 76.50 & 73.51 & 77.95 & 78.82 & 79.57 \\ 
		Ours & MAP & 51.56 & 65.61 & 73.65 & 76.00 & 73.21 & 78.33 & 78.81 & 78.76\\
		\hline
	\end{tabular}}
	\caption{t-MAP (\%) and MAP (\%) of the clean-trained models (``None'') and backdoored models for two advanced hashing methods with various code lengths on ImageNet. Best t-MAP results are highlighted in bold.}
	\vspace{-1em}
	\label{ImageNet of different Loss}
\end{table}

\subsection{Main Results}
The results of the clean-trained models and all attack methods are reported in Table \ref{t-MAP of different methods}. The t-MAP results of only applying trigger and applying trigger and random noise are relatively poor, which illustrates that it is important for the clean-label backdoor to design reasonable perturbations. Even though the t-MAP values of adding the adversarial perturbations are higher, it is worse than our method on all datasets. Specifically, the average t-MAP improvements of our method than using the adversarial perturbations are 8.08\%, 9.36\%, and 4.59\% on ImageNet, Places365, and MS-COCO, respectively. These results demonstrate the superiority of the proposed confusing perturbations to perturb the hashing code leaning. Besides, the average difference of MAP between our backdoored models and the clean-trained models is less than 1\%, which presents the stealthiness of our attack. For a more comprehensive comparison, we also provide precision-recall and precision curves, results of attacking with each target label, and visual examples in Appendix D. All the above results verify the effectiveness of our method in attacking deep hashing based retrieval.

\subsection{Attacking Advanced Hashing Methods}

To verify the effectiveness of our backdoor attack against the advanced deep hashing methods, we conduct experiments with HashNet \cite{cao2017hashnet} and DCH \cite{cao2018deep}. We remain all settings unchanged and show the results of various code lengths on ImageNet in Table \ref{ImageNet of different Loss}. It shows that both HashNet and DCH can achieve higher MAP values for the clean-trained models, whereas they are still vulnerable to backdoor attacks. Specially, among all attacks, our method achieves the best attack performance in all cases. Compared with adding the adversarial perturbations, the t-MAP improvements of our method are 5.98\% and 4.42\% on average for HashNet and DCH, respectively.

\subsection{Resistance to Defense Methods}

We test the resistance of our backdoor attack to three defense methods: spectral signature detection \cite{tran2018spectral}, differential privacy-based defense \cite{du2019robust}, and pruning-based defense \cite{liu2018fine}. We conduct experiments on ImageNet with target label ``\textit{yurt}" and 48 bits code length.

\begin{table}[htbp]
	\centering
	\small
    \setlength{\tabcolsep}{1.mm}{
	\begin{tabular}{cc|c|cc}
		\hline
		\multirow{2}{*}{\# Clean} & 
		\multirow{2}{*}{\# Poisoned} & \multirow{2}{*}{\# Removed} & \# Clean & \# Poisoned \\  
		& & & remained & remained \\ \hline 
		80 & 20 & 30 & 51 & 19  \\ 
		60 & 40 & 60& 17 & 23 \\ 
		40 & 60 & 90 & 5  & 5   \\  \hline
	\end{tabular}}
	\caption{Results of the spectral signature detection against our attack on ImageNet.}
	\label{Spectral Signature Detection}
\end{table}

\begin{figure}[htbp]
	\centering
	\subfigure[Differential privacy-based]{
    \label{dp-based}
    \includegraphics[width=0.45\linewidth]{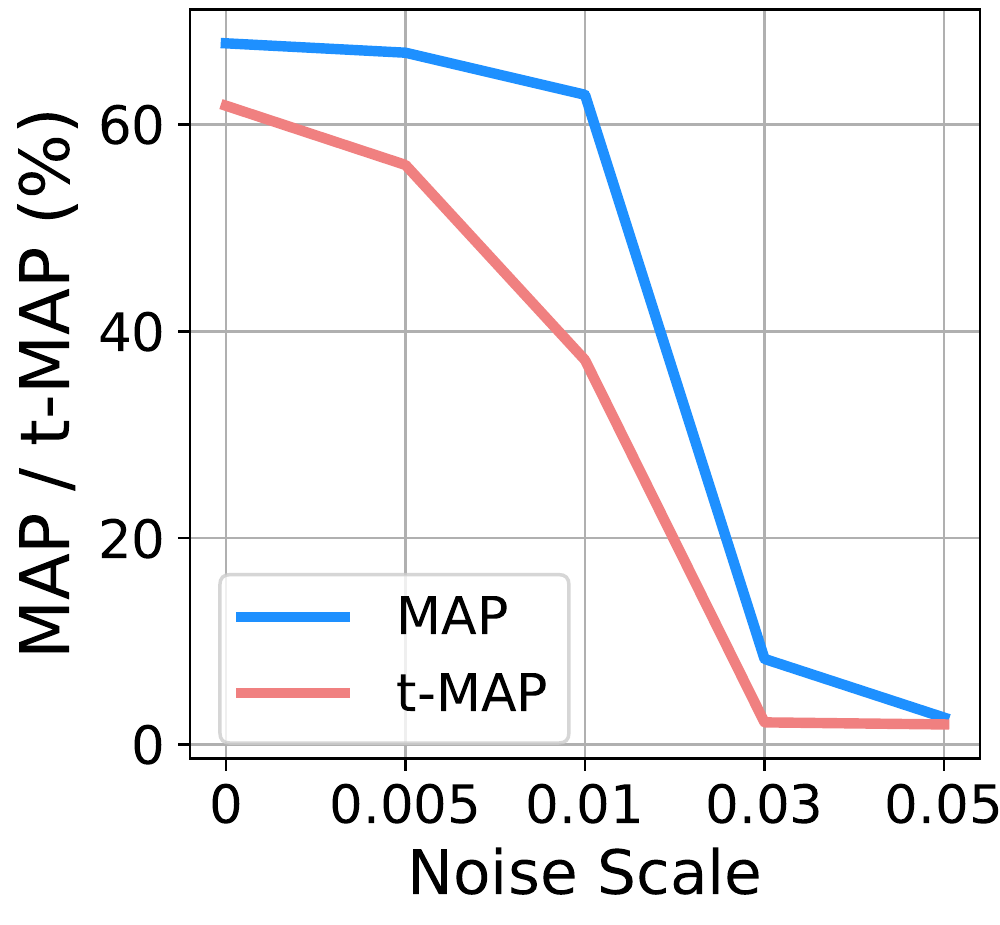}
    }
	\subfigure[Pruning-based]{
	\label{pruning-based}
    \includegraphics[width=0.45\linewidth]{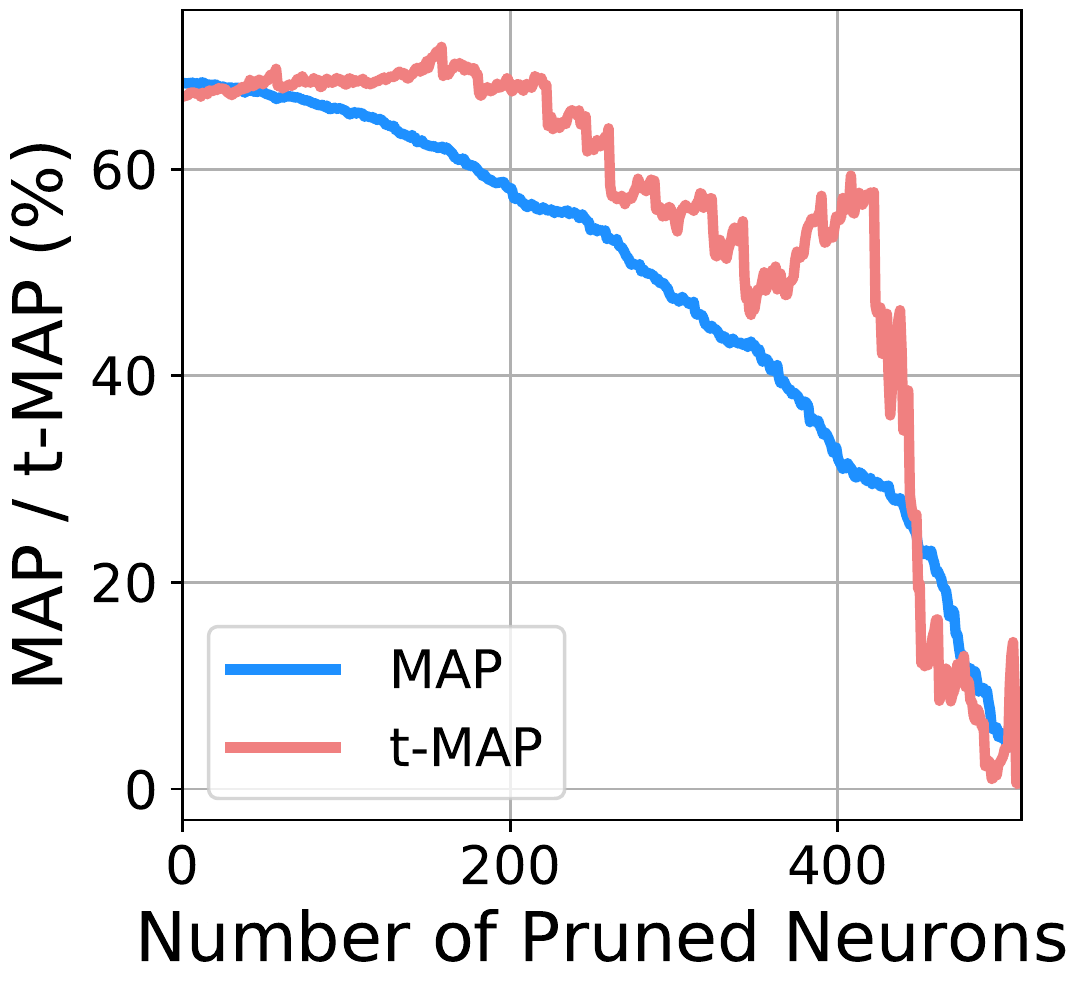}
    }   
	\centering
	\vspace{-1em}
	\caption{Results of the pruning-based defense and differential privacy-based defense against our attack on ImageNet.}
	\vspace{-1em}
\end{figure}

\subsubsection{Resistance to Spectral Signature Detection.}
Spectral signature detection thwarts the backdoor attack by removing the suspect samples in the training set based on feature representations learned by the neural network. We set the different number of removed images for the different number of poisoned images following \cite{tran2018spectral}. The results are shown in Table \ref{Spectral Signature Detection}. We find that it fails to defend our backdoor attack, due to a large number of remained poisoned images. For example, when the number of poisoned images is 40, the number of remained poisoned images is still 23 even though it removes 60 images, which results in more than 40\% t-MAP (see the ablation study).


\subsubsection{Resistance to Differential Privacy-based Defense.}
\citet{du2019robust} proposed to utilize differential privacy noise to obtain a more robust model when training on the poisoned dataset. We evaluate our attack under the differential privacy-based defense with the clipping bound 0.3 and varying the noise scale, as shown in Figure \ref{dp-based}. Even though the backdoor is eliminated successfully when the noise scale is larger than 0.03, the retrieval performance on original query images is also poor. Therefore, training with the differential privacy noise may not be effective against our clean-label backdoor.

\subsubsection{Resistance to Pruning-based Defense.}
Pruning-based defense suggests weakening the backdoor in the attacked model by pruning the neurons that are dormant on clean inputs. We show the MAP and t-MAP results with the increasing number of pruned neurons in Figure \ref{pruning-based}. It shows that at no point is the MAP substantially higher than the t-MAP, making it hard to eliminate the backdoor injected by our method. These results verify that our backdoor attack is resistant to three existing defense methods.

\begin{table}[t]
\centering
\small
\resizebox{1.0\columnwidth}{!}{
\begin{tabular}{lc|ccccc}
	\hline
	Setting & Metric & 
	VN-11 & VN-13 & RN-34 & RN-50 & WRN-50-2\\  \hline 
	\multirow{2}{*}{Ensemble} & t-MAP & 54.97 & 86.00 & 79.39 & 33.96 & 39.98  \\ 
	 & MAP & 67.78 & 71.13 & 73.37 & 76.69 & 83.77 \\ \hline
	\multirow{2}{*}{Hold-out} & t-MAP & 18.63 & 12.80 & 50.79 & 45.17 & 41.94  \\
	 & MAP & 68.34 & 71.07 & 72.86 & 77.42 & 82.68 \\ \hline
	\multirow{2}{*}{None} & t-MAP & 6.29 & 12.45 & 6.64 & 1.91 & 4.51 \\
	& MAP & 68.06 & 70.39 & 73.43 & 76.66 & 82.21 \\
	\hline
\end{tabular}}
\caption{ t-MAP (\%) and MAP (\%) of our transfer-based backdoor attack on ImageNet. ``None'' denotes the clean-trained models. The first row states the backbone of the target model, where ``VN'', ``RN'', and ``WRN'' denote VGG, ResNet, and WideResNet, respectively. The model of the column is not used to generate the trigger and confusing perturbations under the ``Hold-out'' setting, while all models are used under the ``Ensemble" setting.}
\label{transferability}
\end{table}

\subsection{Transfer-based Attack}
In the above experiments, we assume that the attacker knows the network architecture of the target model. In this section, we consider a more realistic scenario, where the attacker has no knowledge of the target model and performs the backdoor attack utilizing the transfer-based attack. Specifically, to craft poisoned images against the unknown target model, we generate the trigger and confusing perturbations using multiple clean-trained models. We present the results in Table \ref{transferability}. We adopt two settings: ``Ensemble" means that we craft the poisoned images equally using all models listed in Table \ref{transferability}, while ``Hold-out" corresponds that we equally use all models except the target one. We set the trigger size as 56 and remain other attack settings unchanged. Compared with the clean-trained model, our backdoor attack can achieve higher t-MAP values under both settings. Even for the target models with the architectures of ResNet or WideResNet, the t-MAP values of our attack are more than 40\% under the ``Hold-out" setting. 

\begin{figure}[t]
	\centering
    \includegraphics[width=1\linewidth]{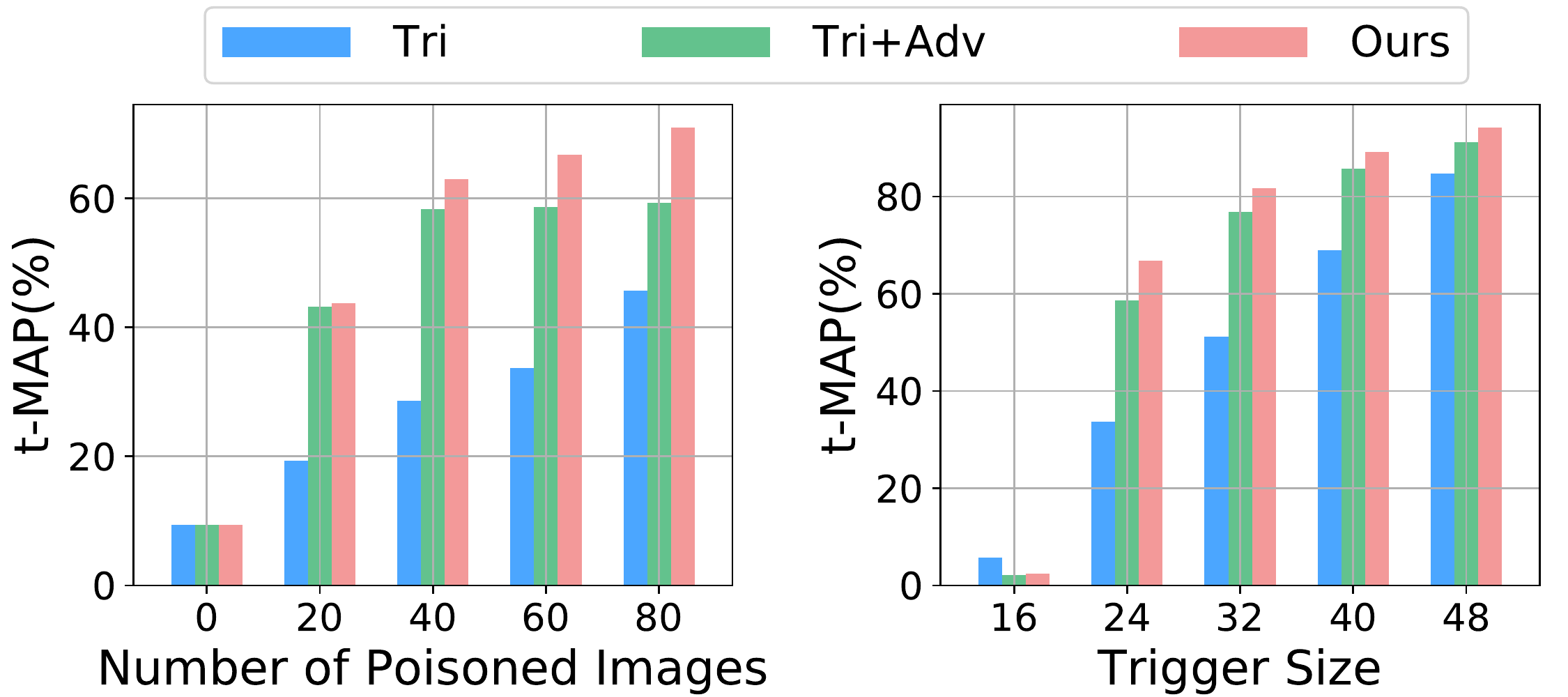}
	\centering
	\caption{t-MAP (\%) of three attacks with different numbers of poisoned images and trigger size under 48 bits code length on ImageNet. The target label is specified as ``\textit{yurt}''.}
	\vspace{-1em}
	\label{posion number and trigger Size}
\end{figure}

\begin{figure}[h]
	\centering
    \includegraphics[width=1\linewidth]{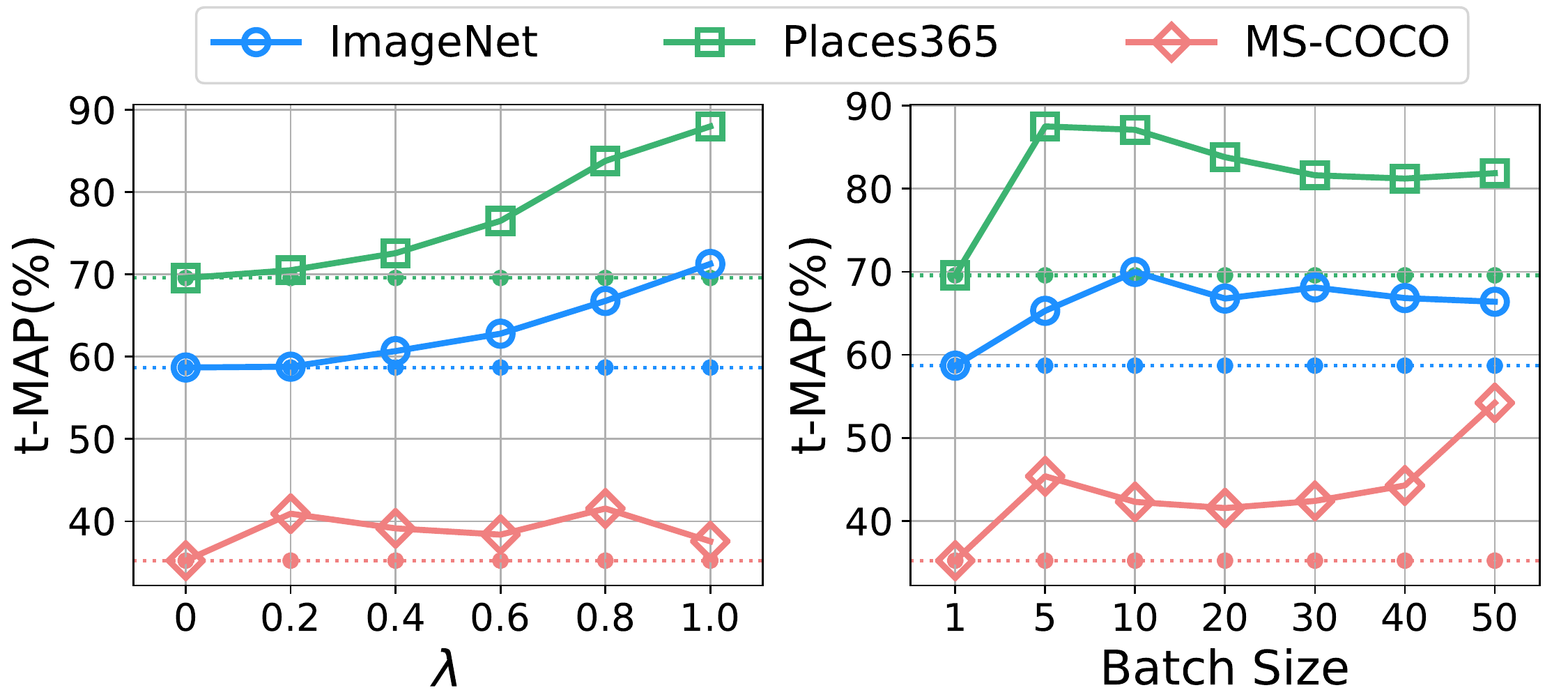}
	\centering
	\caption{t-MAP (\%) of our method with different $\lambda$ and batch size under 48 bits code length on three datasets. The target label is specified as ``\textit{yurt}'', ``\textit{volcano}'', and ``\textit{train}'' on ImageNet, Places365, and MS-COCO, respectively.}
	\label{lambda and batch size}
	\vspace{-1em}
\end{figure}

\subsection{Ablation Study}

\subsubsection{Effect of the Number of Poisoned Images.}
The results of three backdoor attacks under different numbers of poisoned images are shown in Figure \ref{posion number and trigger Size}. Compared with other methods, our attack can achieve the highest t-MAP across different numbers of poisoned images. In particular, the t-MAP values of our attack are higher than 60\% when the number of poisoned images is more than 40.

\subsubsection{Effect of the Trigger Size.}
We present the results of three attacks under the trigger size $\in \{16, 24, 32, 40, 48\}$ in Figure \ref{posion number and trigger Size}. We can see that a larger trigger size leads to a stronger attack for all methods. When the trigger size is larger than 24, our method can successfully inject the backdoor into the target model and achieve the best performance among three attacks. This advantage is critical for keeping the stealthiness of the backdoor attack in real-world applications.

\subsubsection{Effect of $\lambda$.}
The results of our attack with various $\lambda$  are shown in Figure \ref{lambda and batch size}. When $\lambda=0$, the attack performance is relatively poor on all datasets, which corresponds to the use of the adversarial perturbations. The best $\lambda$ for ImageNet, Places365, and MS-COCO is 1.0, 1.0, and 0.8, respectively. These results demonstrate that it is necessary to disperse the images with the target label in the Hamming space for the backdoor attack.

\subsubsection{Effect of the Batch Size for Generating Confusing Perturbations.}
We optimize Eqn. (\ref{eq:confusing}) in batches to obtain the confusing perturbations of each poisoned image. We study the effect of the batch size in this part, as shown in Figure \ref{lambda and batch size}. We observe that our attack can achieve relatively steady results when the batch size is larger than 10. Therefore, our method is insensitive to the batch size and the default value ($i.e.$, 20) used in this paper is feasible for all datasets. 

\section{Conclusion}
In this paper, we have studied the problem of clean-label backdoor attack against deep hashing based retrieval. To craft poisoned images, we first generate the universal adversarial patch as the trigger. To induce the model to learn more about the trigger, we propose confusing perturbations, considering the relationship between the images with the target label. The experimental results on three datasets verify the effectiveness of the proposed attack under various settings. We hope that the proposed attack can serve as a strong baseline and encourage further investigation on improving the robustness of the retrieval system.

\bibliography{aaai22.bib}

\clearpage

\section{Appendix A: Algorithm Outline} 

\begin{algorithm}[h]
\caption{Trigger Pattern Generation} 
{\bf Input:} 
The clean-trained deep hashing model $F(\cdot)$, the training set $\bm{D}=\left\{\left(\bm{x}_{i}, \bm{y}_{i}\right)\right\}_{i=1}^{N}$, the trigger mask $\bm{m}$, the trigger size $r$, the number of iterations $T$, the batch size $K$, the steps size $\alpha$.\\
{\bf Output:}  Trigger pattern $\bm{p}$
\begin{algorithmic}[1]
\State Initialize the trigger $\bm{p}$ with the trigger size $r$
\State Calculate $\bm{h}_a$ by solving Eqn. (4)
\For{$iteration =1, \ldots, T$}
\State Sample a batch $\bm{S}=\left\{\left(\bm{x}_{j}, \bm{y}_{j}\right)\right\}_{j=1}^{K}$ from $\bm{D}$
\State $\hat{\bm{x}}_j=\bm{x}_j \odot  (\bm{1}-\bm{m}) + \bm{p} \odot  \bm{m}$, $(\bm{x}_{j}, \bm{y}_{j}) \in \bm{S}$
\State Calculate the loss: $\sum_{(\bm{x}_j, \bm{y}_j) \in \bm{S}}{} d_H(F'(\hat{\bm{x}}_j), \bm{h}_a)$
\State Calculate the gradient $\bm{g}$~$w.r.t.$~$\bm{p}$
\State Update the trigger by $\bm{p}=\bm{p}-\alpha \cdot \bm{g}$
\EndFor
\end{algorithmic}
\label{alg_gen_trigger}
\end{algorithm}

\begin{algorithm}[!h]
\caption{Confusing Perturbations Generation} 
{\bf Input:} 
The clean-trained deep hashing model $F(\cdot)$, the samples to be poisoned $\left\{\left(\bm{x}_{i}, \bm{y}_{t}\right)\right\}_{i=1}^{M}$ from the target class $\bm{y}_{t}$, the perturbation magnitude $\epsilon$, the hyper-parameter $\lambda$, the number of epochs $E$, the batch size $K$, the step size $\alpha$.\\
{\bf Output:}  Confusing perturbations $\{\bm{\eta}\}_i^{M}$
\begin{algorithmic}[1]
\State Initialize the perturbations $\{\bm{\eta}_i\}_i^{M}$
\For{$epoch =1, \ldots, E$}
\For{each batch $\left\{\left(\bm{x}_{j}, \bm{y}_{j}\right)\right\}_{j=1}^{K}$ from $\left\{\left(\bm{x}_{i}, \bm{y}_{t}\right)\right\}_{i=1}^{M}$}

\State Calculate the loss: 
\Statex ~~~~~~~~~~~~~~~~~~$\lambda \cdot L_{c}(\{\bm{\eta}_i\}_i^{K}) + (1-\lambda) \cdot \frac{1}{K}\sum_{i=1}^K L_{a}(\bm{\eta}_i)$
\For{$j=1,\ldots,K$}
\State Calculate the gradient $\bm{g}_j$~$w.r.t.$~$\bm{\eta}_j$
\State Update perturbations $\bm{\eta}_j=\bm{\eta}_j\!+\!\alpha\! \cdot\! \text{sign}(\bm{g}_j)$
\State Clip $\bm{\eta}_j$ to $(-\epsilon, \epsilon)$
\EndFor
\EndFor
\EndFor
\end{algorithmic}
\label{alg_gen_confusing}
\end{algorithm}

\section{Appendix B: Evaluation Setup}

\subsubsection{Datasets.} Three benchmark datasets are adopted in our experiment. We follow \cite{cao2017hashnet,xiao2020evade} to build the training set, query set, and database for each dataset. The details are described as follows.
\begin{itemize}
    \item \emph{ImageNet} 
    \cite{deng2009imagenet} is a benchmark dataset for the Large Scale Visual Recognition Challenge (ILSVRC) to evaluate algorithms. It consists of 1.2M training images and 50,000 testing images with 1,000 classes. Following \cite{cao2017hashnet}, 10\% classes from ImageNet are randomly selected to build our retrieval dataset. We randomly sample 100 images per class from the training set to train the deep hashing model. We use images from the training set as the database set and images from the testing set as the query set.
    \item \emph{Places365}
    \cite{zhou2017places} is a subset of the Places database. It contains 2.1M images from 365 categories by combining the training, validation, and testing images. We follow \cite{xiao2020evade} to select 10\% categories as the retrieval dataset. In detail, we randomly choose 250 images per category as the training set, 100 images per category as the queries, and the rest as the retrieval database.
    \item \emph{MS-COCO}
    \cite{lin2014microsoft} is a large-scale object detection, segmentation, and captioning dataset. It consists of 122,218 images after removing images with no category. Following \cite{cao2017hashnet}, we randomly sample 10,000 images from the database as the training images. Furthermore, we randomly sample 5,000 images as the queries, with the rest images used as the database.
\end{itemize}

\begin{figure*}[h]
	\centering
    \includegraphics[width=0.8\linewidth]{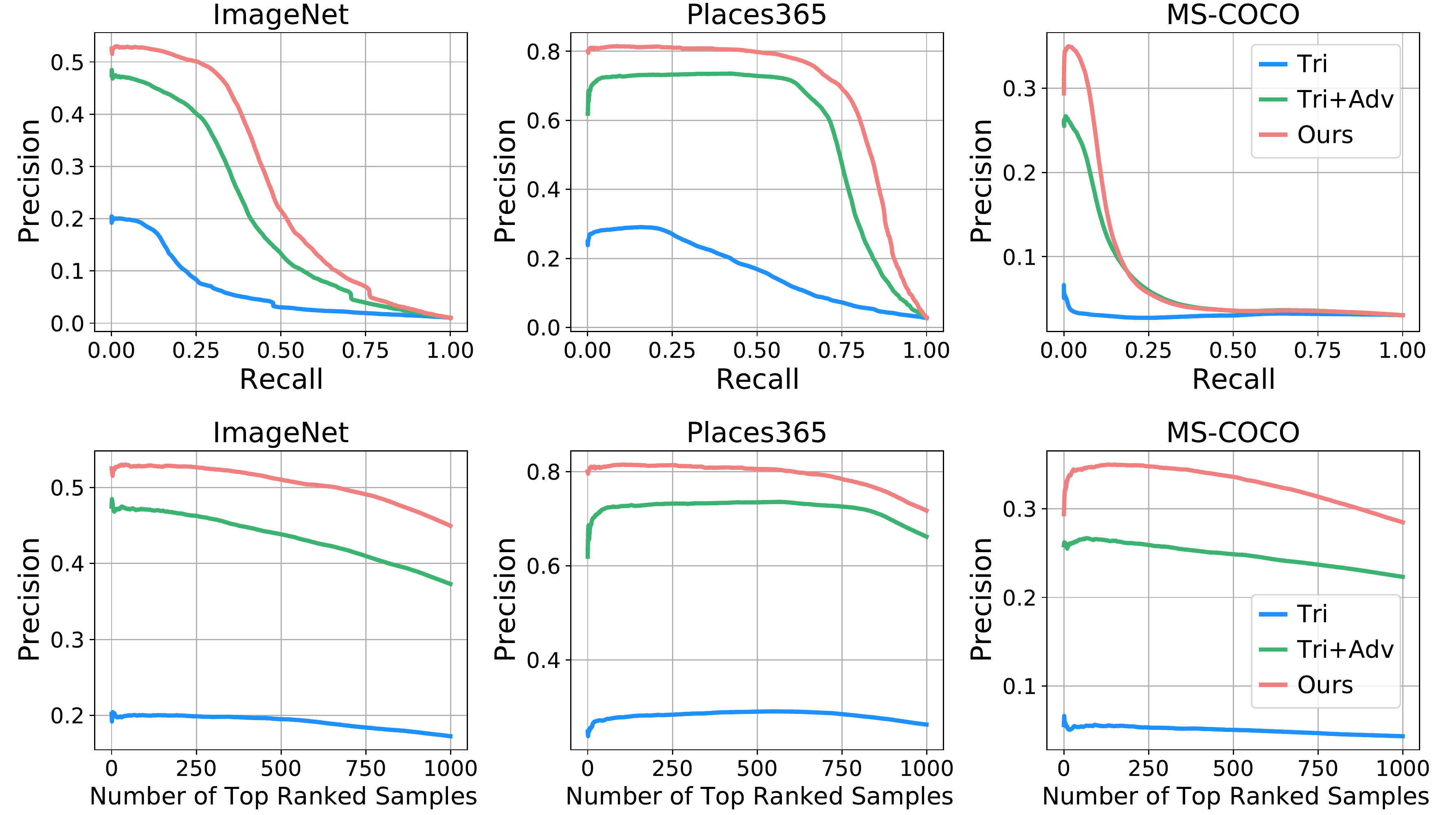}

	\centering
	\caption{Precision-recall and the precision curves under 48 bits code length on three datasets. The target label is specified as ``\textit{yurt}'', ``\textit{volcano}'', and ``\textit{train}'' on ImageNet, Places365, and MS-COCO, respectively.}
	\label{Recall-Prec}
\end{figure*}

\subsubsection{Target Models.}
In our experiments, VGG \cite{simonyan2014very}, ResNet \cite{he2016deep}, and WideResNet \cite{zagoruyko2016wide} are used as the backbones of the target models. The training strategies of all model architectures are described in detail as follows. Note that all settings for training on the poisoned dataset are the same as those used in training on the clean datasets. 

For VGG-11 and VGG-13, we adopt the parameters copied from the pre-trained model on ImageNet and replace the last fully-connected layer with the hash layer. Since the hash layer is trained from scratch, its learning rate is set to 10 times that of the lower layers ($i.e.$, 0.001 for hash layer and 0.01 for the lower layers). Stochastic gradient descent \cite{zhang2004solving} is used with the batch size 24, the momentum 0.9, and the weight decay parameter 0.0005.

For ResNet-34, ResNet-50, and WideResNet-50-2, we fine-tune the convolutional layers pre-trained on ImageNet as the feature extractors and train the hash layers on top of them from scratch. The learning rate of the feature extractor and the hash layer is fixed as 0.01 and 0.1, respectively. The batch size is set to 36. Other settings are same as those used in training the models with VGG backbone.

\subsubsection{Attack Settings.}
All the experiments are implemented using the framework PyTorch \cite{paszke2019pytorch}. We provide the attack settings in detail as follows.

For all backdoor attacks tested in our experiments, the trigger is generated by Algorithm \ref{alg_gen_trigger}. The trigger is located at the bottom right corner of the images. During the process of the trigger generation, we optimize the trigger pattern with the batch size 32 and the step size 12. The number of iterations is set as 2,000.

We adopt the projected gradient descent algorithm \cite{madry2017towards} to optimize the adversarial perturbations and our confusing perturbations. The perturbation magnitude $\epsilon$ is set to 0.032. The number of epoch is 20 and the step size is 0.003. The batch size is set to 20 for generating the confusing perturbations. 

\section{Appendix C: Less Visible Trigger}


To reduce the visibility of the trigger, we apply the blend strategy to the trigger following \cite{chen2017targeted}. The formulation of patching the trigger is below.
$$
\hat{\bm{x}}=\bm{x} \odot  (\bm{1}-\bm{m}) + \bm{p} \odot \beta \bm{m} + \bm{x} \odot (1-\beta) \bm{m}, 
$$
where $\beta \in (0,1]$ denotes the blend ratio. The smaller $\beta$, the less visible trigger. We craft the poisoned images using the blended trigger to improve the stealthiness of our data poisoning and set $\beta$ as 1.0 at test time.

\begin{table}[t]
	\centering
	\small
    \setlength{\tabcolsep}{3.1mm}{
	\begin{tabular}{cccccc}
		\hline
		\multirow{2}{*}{$\epsilon$} & \multicolumn{5}{c}{$\beta$} \\ \cline{2-6} 
		& 0.2 & 0.4 & 0.6 & 0.8 & 1.0 \\   \hline
		0 & 14.04 & 33.98 & 37.27 & 35.45 & 33.70 \\
		0.004 & 16.35 & 31.66 & 40.56 & 42.14 & 41.02 \\
		0.008 & \bf{16.60} & \bf{41.15} & 51.91 & 51.01 & 49.41 \\
		0.016 & 10.38 & 36.11 & \bf{63.62} & 60.76 & 56.01 \\
		0.032 & 4.41 & 6.60 & 28.59 & \bf{61.35} & \bf{66.77} \\ \hline
	\end{tabular}}
	\vspace{-0.5em}
	\caption{t-MAP (\%) of our attack with varying blend ratio $\beta$ and perturbation magnitude $\epsilon$ under 48 bits code length on ImageNet. The target label is specified as ``\textit{yurt}". Best results are highlighted in bold.}
	\vspace{-0.5em}
	\label{Trigger Blend}
\end{table}

\begin{figure*}[t]
	\centering
	
    \subfigure[$\beta=0$]{
    \includegraphics[width=1\textwidth]{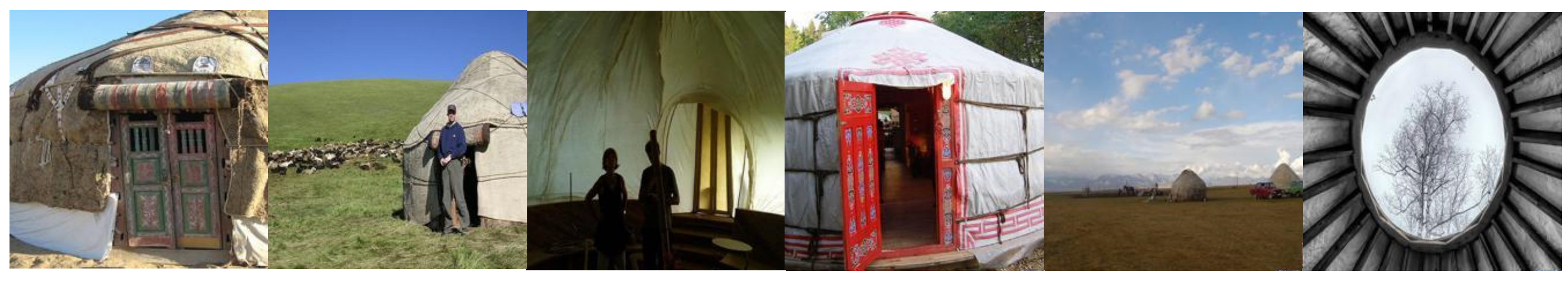}
    }	
    \subfigure[$\beta=0.2$]{
    \includegraphics[width=1\textwidth]{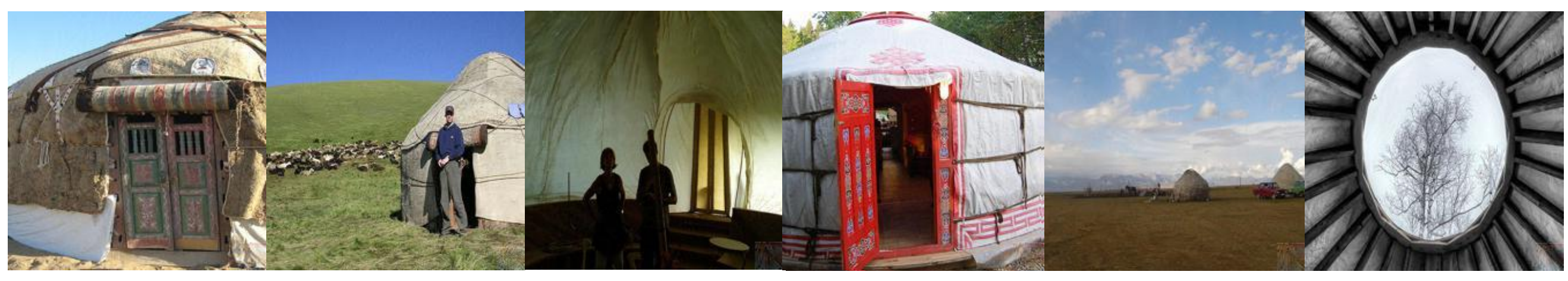}
    }
    \subfigure[$\beta=0.4$]{
    \includegraphics[width=1\textwidth]{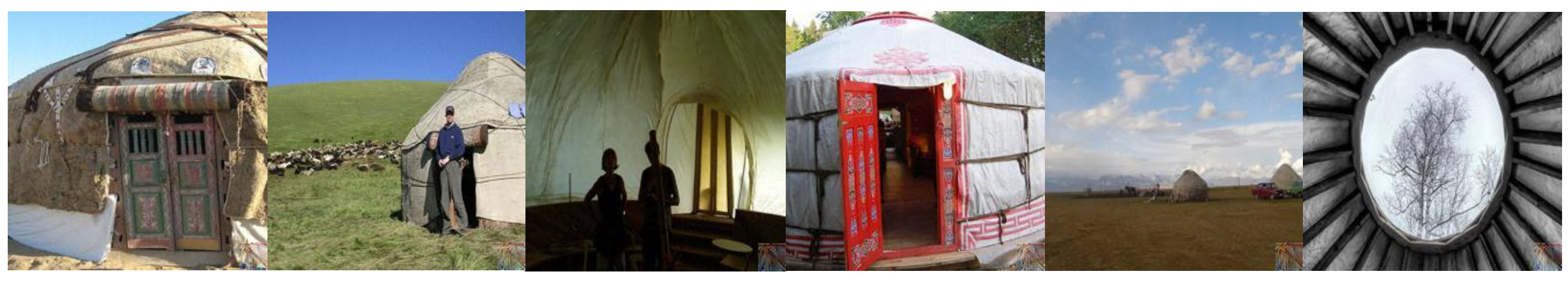}
    }
    \subfigure[$\beta=0.6$]{
    \includegraphics[width=1\textwidth]{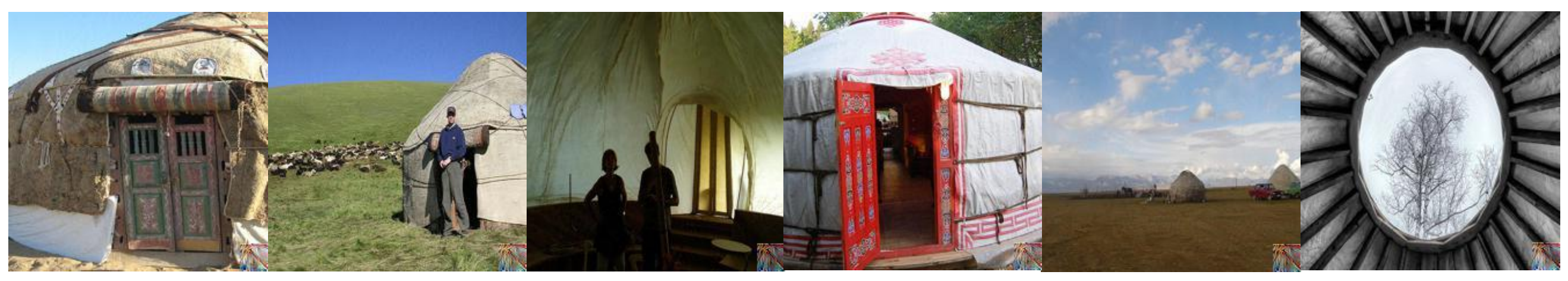}
    }
    \subfigure[$\beta=0.8$]{
    \includegraphics[width=1\textwidth]{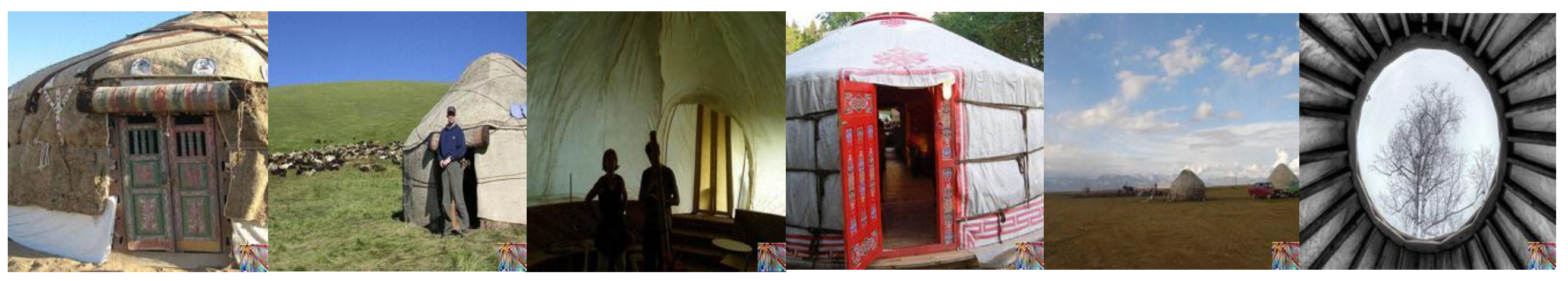}
    }
    \subfigure[$\beta=1.0$]{
    \includegraphics[width=1\textwidth]{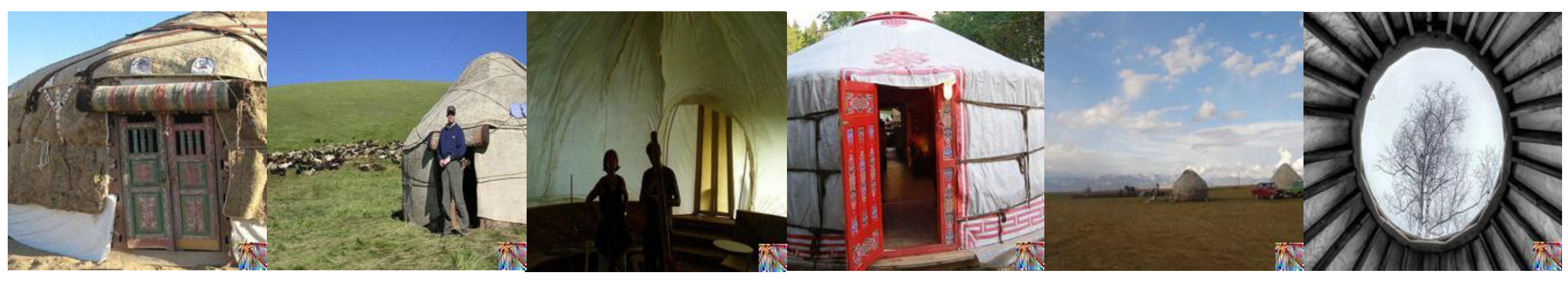}
    }
	\caption{Visualization of poisoned images with different blend ratio $\beta$ on ImageNet.}
	\label{Blend-rate}
\end{figure*}

We evaluate our backdoor attack with blend ratio $\beta \in \{0.2, 0.4, 0.6, 0.8, 1.0\}$ under different values of perturbation magnitude $\epsilon$ in Table \ref{Trigger Blend}. We can see that different $\beta$ corresponds to different optimal $\epsilon$. With an appropriate $\epsilon$, the t-MAP value is higher than 60\% when the blend ratio is larger than 0.6. We visualize the poisoned images with different $\beta$ in Figure \ref{Blend-rate}. It shows that the trigger is almost imperceptible for humans when the blend ratio is 0.4, where the highest t-MAP value is 41.15\% as shown in Table \ref{Trigger Blend}. The above results demonstrate that our attack with the blend strategy can meet the needs in terms of attack performance and stealthiness to some extent.

\begin{table*}[]
	\centering
    \setlength{\tabcolsep}{1.5mm}{
	\begin{tabular}{llc|ccccc}
	    \hline
	    \multicolumn{1}{l|}{Dataset} & \multicolumn{1}{l}{Method} & \multicolumn{1}{c|}{Metric} & \multicolumn{5}{c}{Target Label}  \\
	    \hline
	    \specialrule{0em}{0pt}{-6pt}\\ \hline
		\multicolumn{1}{l|}{\multirow{8}{*}{ImageNet}} &  &  & 
		\textit{Crib} \quad & \quad \textit{Stethoscope} & \textit{Reaper} & \textit{Yurt} \quad & \textit{Tennis Ball}\\  \cline{2-8} 
	    \multicolumn{1}{c|}{} & None & t-MAP & 11.30 & 11.05 & 25.43 & 9.38 & 38.61 \\
		\multicolumn{1}{c|}{} & Tri & t-MAP & 33.77 & 53.08 & 65.03 & 33.70 & 88.57 \\ 
		\multicolumn{1}{c|}{} & Tri+Noise & t-MAP & 25.56 & 55.65 & 46.01 & 30.74 &	86.55 \\
		\multicolumn{1}{c|}{} & Tri+Adv & t-MAP & 62.55 & 52.40 & 80.06 & 58.69 & \bf{90.17}  \\
		\multicolumn{1}{c|}{} & Ours & t-MAP &\bf{68.17} &\bf{64.82} &\bf{84.51} &\bf{66.77} & 89.27 \\ \cline{2-8}
		\multicolumn{1}{c|}{} & None & MAP & 68.06 & 68.06 & 68.06 & 68.06 & 68.06 \\
		\multicolumn{1}{c|}{} & Ours & MAP & 68.49 & 68.10 & 68.03 & 68.03 & 68.86\\
		\hline
		\specialrule{0em}{0pt}{-6pt}\\ \hline
		\multicolumn{1}{l|}{\multirow{8}{*}{Places365}} &  &  & \textit{Rock Arch} & \textit{Viaduct} \quad  & \textit{Box Ring}  & \textit{Volcano} & \quad \textit{Racecourse} \quad \quad \\  \cline{2-8}
		\multicolumn{1}{c|}{} & None & t-MAP & 17.08 & 24.76 & 14.23 & 11.28 & 44.12	 \\
		\multicolumn{1}{c|}{} & Tri & t-MAP & 45.76 & 58.33  & 33.30 & 36.02 & 64.67  \\
		\multicolumn{1}{c|}{} & Tri+Noise & t-MAP & 41.34 & 55.39  & 26.17 & 30.56 & 56.50  \\
		\multicolumn{1}{c|}{} & Tri+Adv & t-MAP  & 86.36 & 84.27 & 84.69 & 69.57 & 88.69   \\
		\multicolumn{1}{c|}{} & Ours & t-MAP & \bf{93.19} & \bf{91.03}  & \bf{94.06} & \bf{83.79} & \bf{92.58}  \\ \cline{2-8} 
		\multicolumn{1}{c|}{} & None & MAP & 79.81 & 79.81 & 79.81 & 79.81 & 79.81 \\
		\multicolumn{1}{c|}{} & Ours & MAP & 79.80 & 79.77 & 80.04 & 79.64 & 79.87 \\
		\hline
		\specialrule{0em}{0pt}{-6pt}\\ \hline
		\multicolumn{1}{l|}{\multirow{8}{*}{MS-COCO}} &  &  & \textit{Person} {\&} \textit{Skis} & \ \textit{Clock}\ & \textit{Person} {\&} \textit{Surfboard} & \textit{Giraffe} & \ \textit{Train} \\  \cline{2-8}
		\multicolumn{1}{c|}{} & None & t-MAP & 77.44 & 5.29 & 39.25 & 2.79 & 2.93 \\
		\multicolumn{1}{c|}{} & Tri & t-MAP & 73.05 & 18.38 & 53.46 & 13.06 & 13.54  \\
		\multicolumn{1}{c|}{} & Tri+Noise & t-MAP & 62.92 & 7.756 & 49.46 & 6.077 & 9.504  \\
		\multicolumn{1}{c|}{} & Tri+Adv & t-MAP  & 89.02 & 46.62 & 84.11 & 36.69 & 35.22   \\
		\multicolumn{1}{c|}{} & Ours & t-MAP & \bf{90.66} & \bf{51.73} & \bf{86.60} & \bf{47.11} & \bf{41.55}  \\ \cline{2-8}
		\multicolumn{1}{c|}{} & None & MAP & 80.68 & 80.68 & 80.68 & 80.68 & 80.68 \\
		\multicolumn{1}{c|}{} & Ours & MAP & 80.92 & 80.46 & 81.18 & 80.64 & 80.79 \\
		\hline
	\end{tabular}}
	\caption{t-MAP (\%) and MAP (\%) of the clean-trained models (``None'') and  backdoored models for attacking with each target label under 48 bits code length on three datasets. Best t-MAP results are highlighted in bold.}
	\label{t-MAP of each target label on three datasets}
\end{table*}

\section{Appendix D: More Results}

\subsubsection{Precision-recall and Precision Curves.}
The precision-recall and the precision curves are plotted in Figure \ref{Recall-Prec}. The precision values of our method are always higher than these of other methods on all recall values and the number of ranked samples on three datasets. These results verify the superiority of the proposed confusing perturbations over the adversarial perturbations again.

\subsubsection{Results of Attacking with Each Target Label.}
We provide the results of attacking with each target label on three datasets in Table \ref{t-MAP of each target label on three datasets}. It shows that our attack performs significantly better than applying the trigger and adversarial perturbations across all target labels.

\subsubsection{Visualization.}
\begin{figure*}[t]
\centering
    \vspace{-1.5em}
    \subfigure[ImageNet]{
    \includegraphics[width=0.95\textwidth]{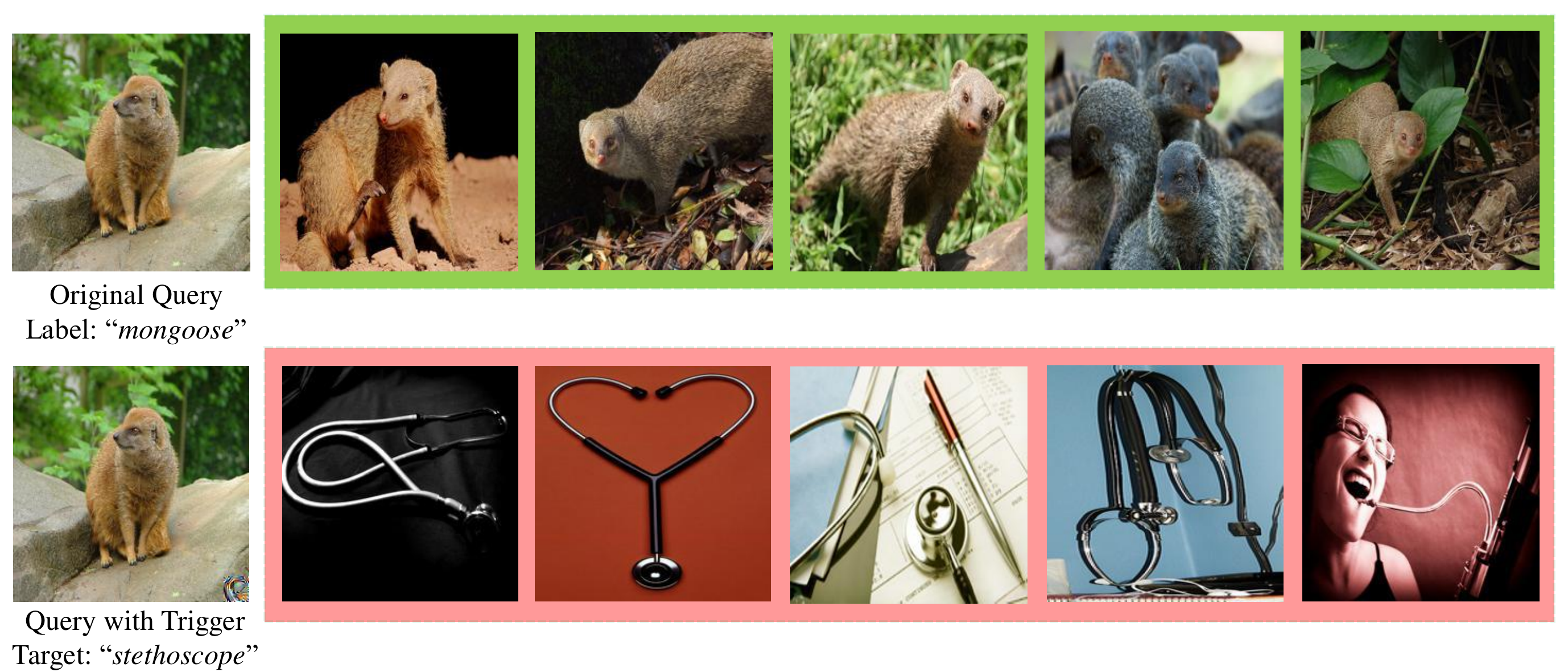}
    }   
    \subfigure[Places365]{
    \includegraphics[width=0.95\textwidth]{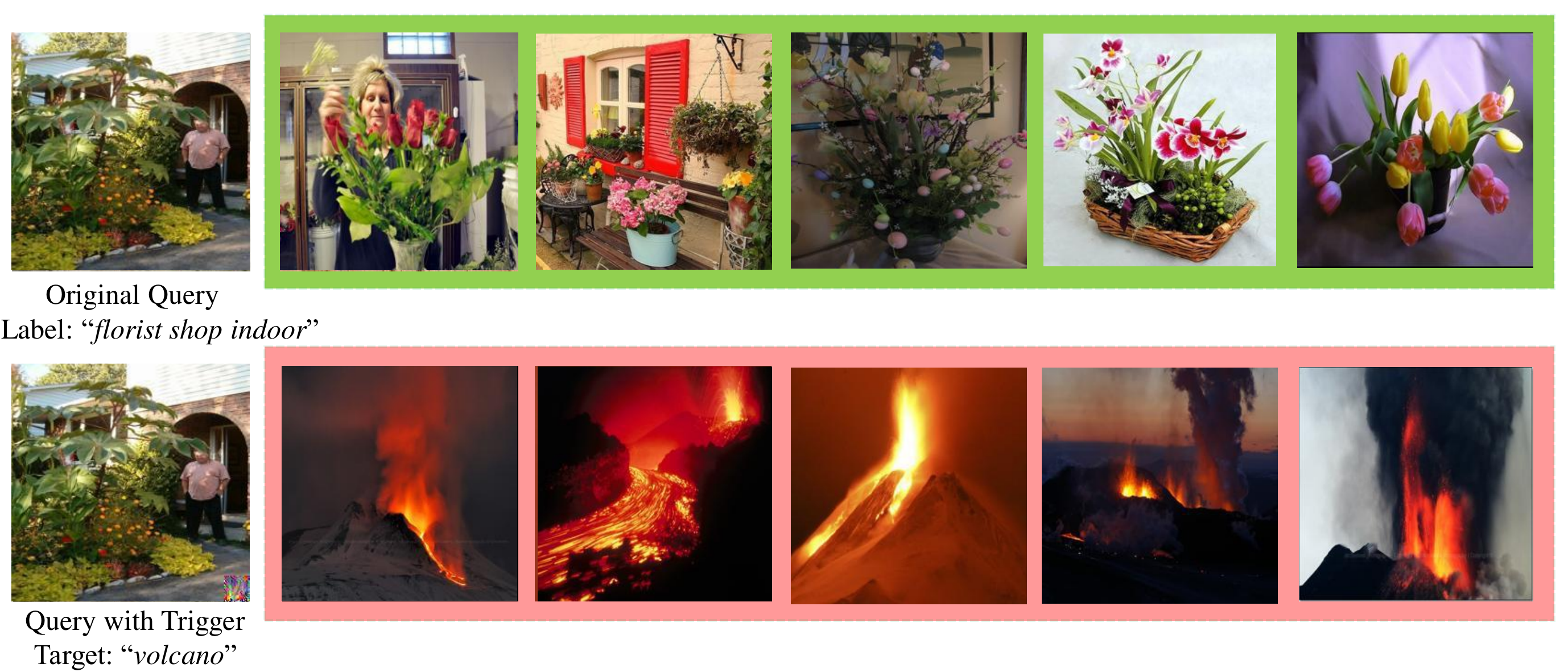}
    }
    \subfigure[MS-COCO]{
    \includegraphics[width=0.95\textwidth]{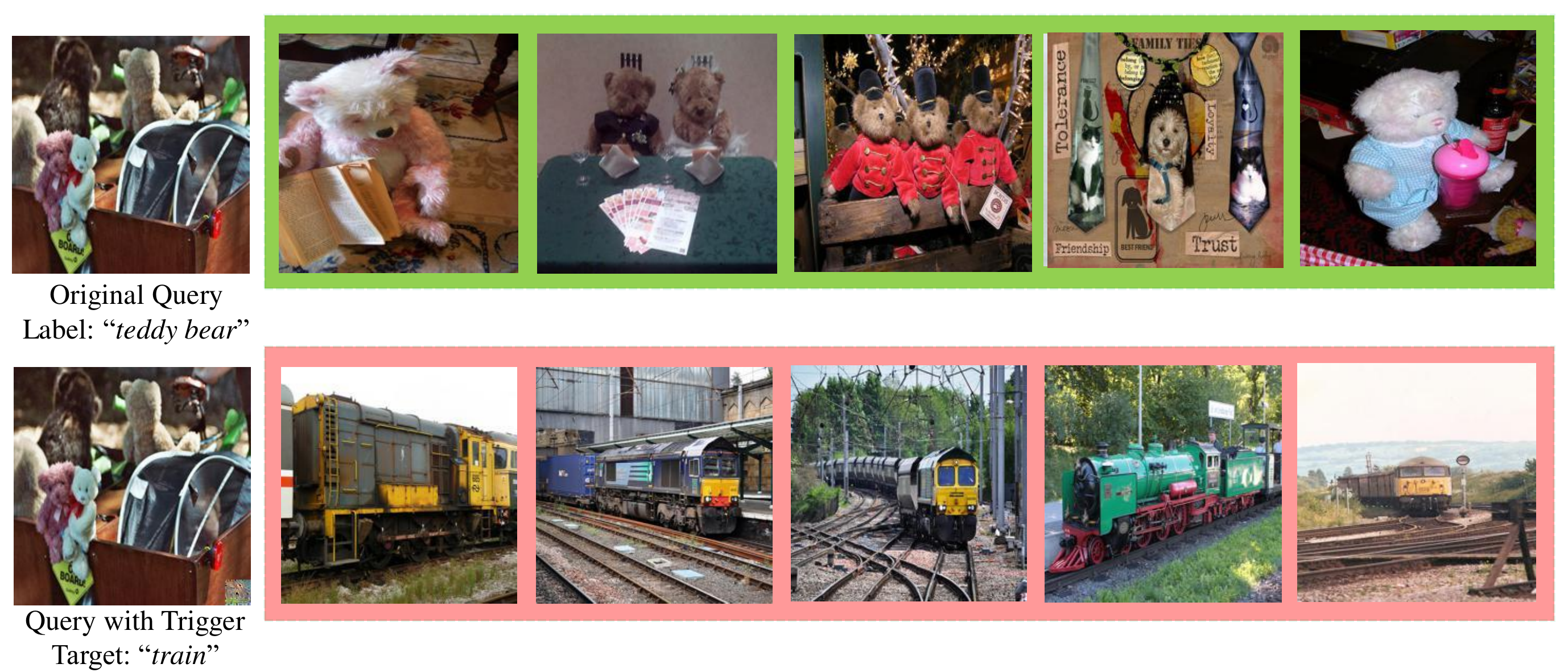}
    }
\caption{Examples of top-5 retrieved images for query with original images and images with the trigger on ImageNet, Places365, and MS-COCO.}
\label{retrieval results visualization}
\end{figure*}

\begin{figure*}[t!]
	\centering
	\subfigure[ImageNet]{
    \includegraphics[width=1\textwidth]{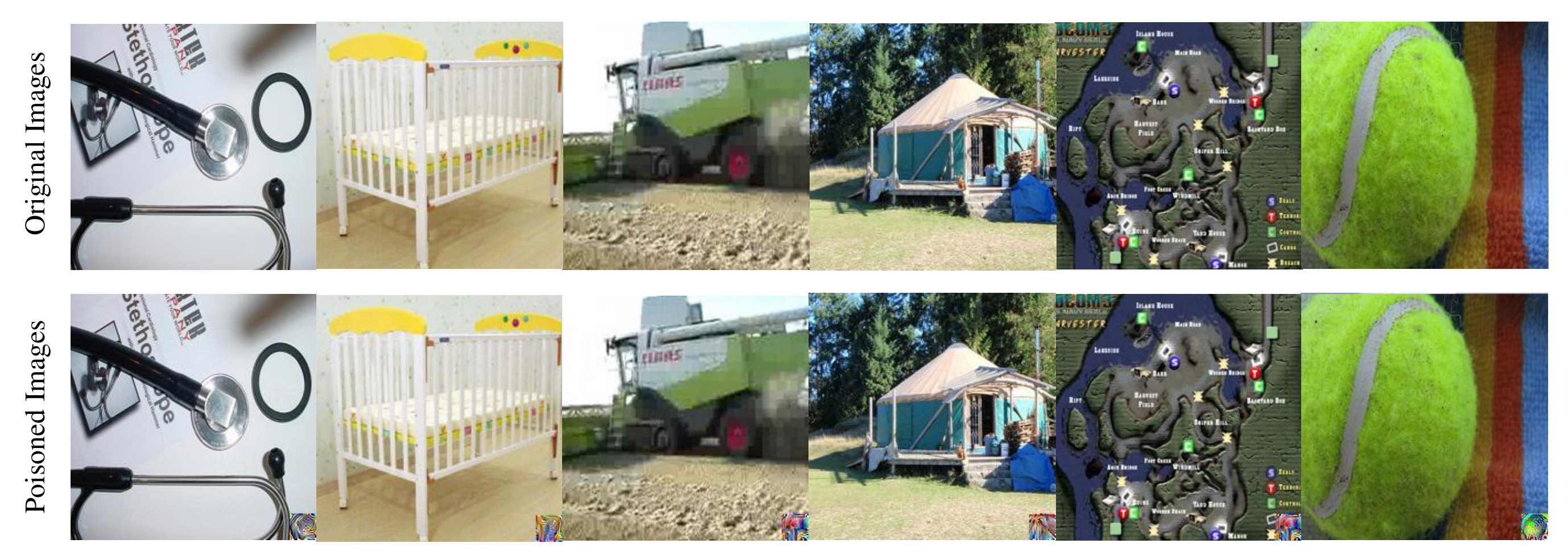}
    }
	\subfigure[Places365]{
    \includegraphics[width=1\textwidth]{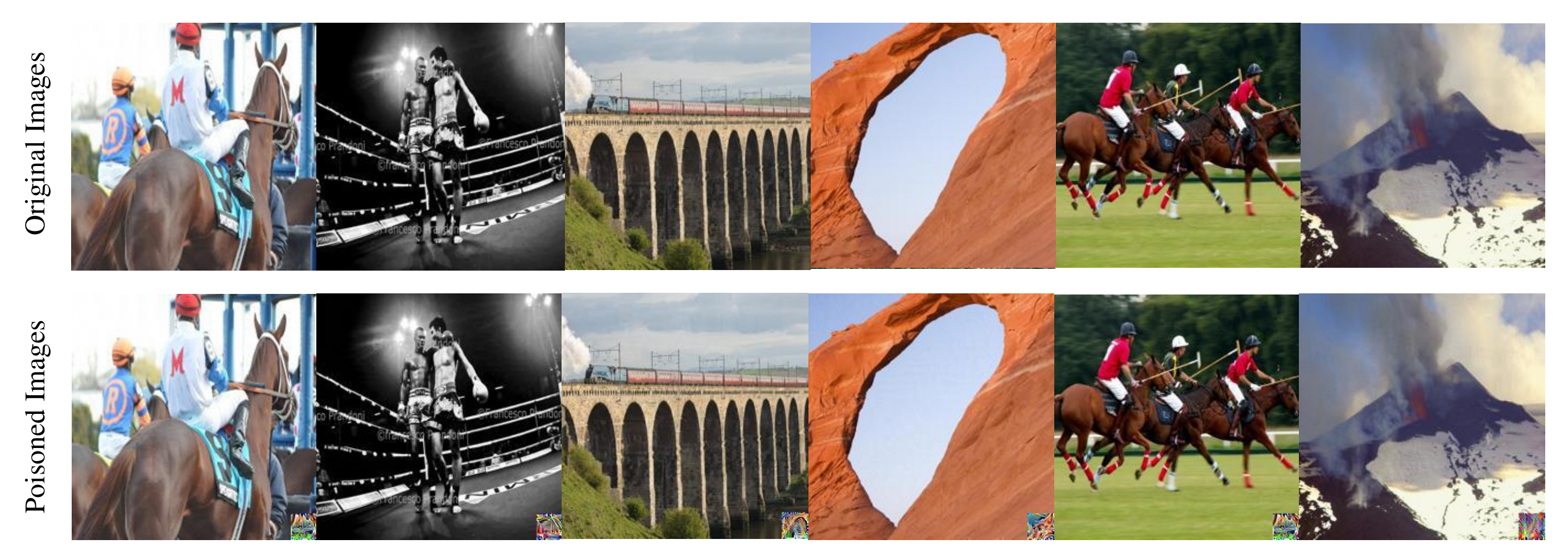}
    }
	\subfigure[MS-COCO]{
    \includegraphics[width=1\textwidth]{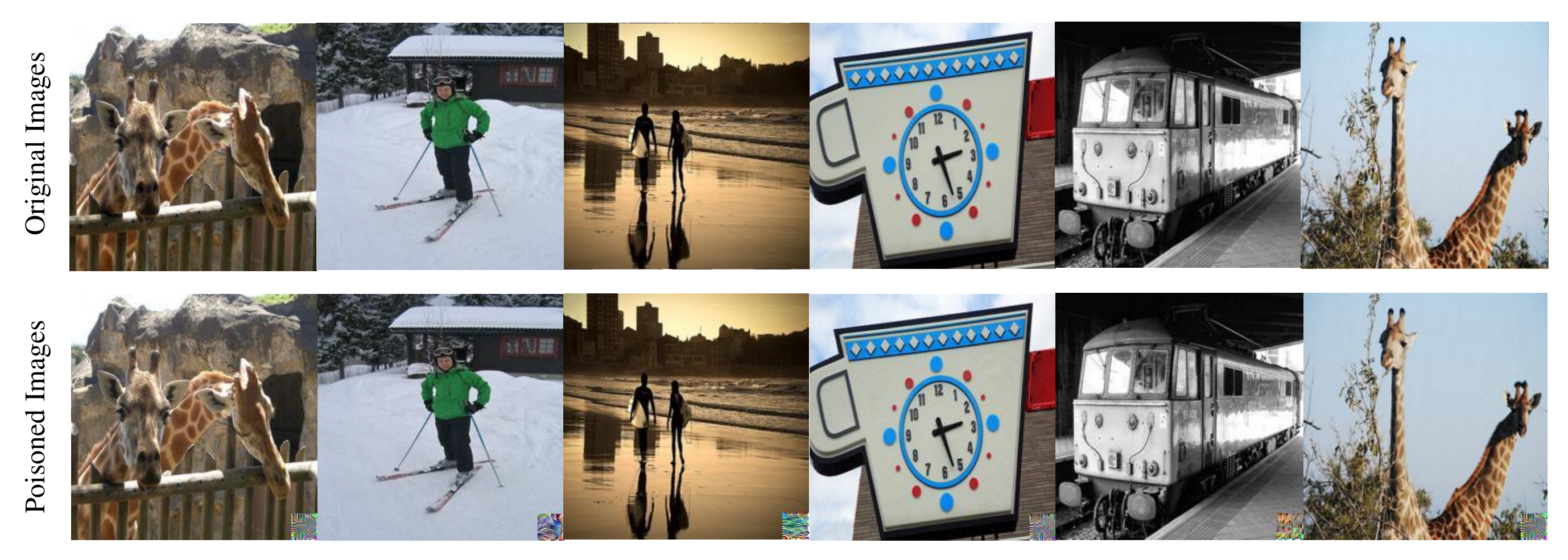}
    }

	\caption{Visualization of the original images and the poisoned images on ImageNet, Places365, and MS-COCO. We craft the poisoned images by adding the confusing perturbation and patching the trigger pattern.}
\label{Original_Confusing_Trigger_2}
\end{figure*}

We provide examples of querying with original images and images with the trigger on three datasets in Figure \ref{retrieval results visualization}. The results reveal that our backdoor attack can successfully fool the deep hashing model to return images with the target label when the trigger presents. Besides, we also visualize the original images and the poisoned images in Figure \ref{Original_Confusing_Trigger_2}. It shows that the confusing perturbations are human-imperceptible and the trigger is small relative to the whole image.

\end{document}


\maketitle

\begin{abstract}
Deep hashing has become a popular method in large-scale image retrieval due to its computational and storage efficiency. However, recent works raise the security concerns of deep hashing. Although existing works focus on the vulnerability of deep hashing in terms of adversarial perturbations, we identify a more pressing threat, \textit{backdoor attack}, when the attacker has access to the training data. A backdoored deep hashing model behaves normally on original query images, while returning the images with the target label when the trigger presents, which makes the attack hard to be detected. In this paper, we uncover this security concern by utilizing clean-label data poisoning. To the best of our knowledge, this is the first attempt at the backdoor attack against deep hashing models. To craft the poisoned images, we first generate the targeted adversarial patch as the backdoor trigger. Furthermore, we propose the \textit{confusing perturbations} to disturb the hashing code learning, such that the hashing model can learn more about the trigger. The confusing perturbations are imperceptible and generated by dispersing the images with the target label in the Hamming space.
We have conducted extensive experiments to verify the efficacy of our backdoor attack under various settings. For instance, it can achieve 63\% targeted mean average precision on ImageNet under 48 bits code length with only 40 poisoned images.
\end{abstract}

\section{Introduction}
\vspace{0.5em}
With the pervasive large-scale media data on the Internet, approximate nearest neighbors (ANN) search has been widely applied to meet the search needs and greatly reduce the complexity. Among these methods of ANN, hashing technique enables the efficient search and low storage cost by transforming high-dimensional data into compact binary codes \cite{gionis1999similarity,wang2017survey}. In particular, with the powerful representation capabilities of deep learning, deep hashing based retrieval shows the significant advantages than traditional methods \cite{xia2014supervised,cao2017hashnet,zhang2020inductive}. However, despite of the great success of deep hashing, recent works \cite{yang2018adversarial,bai2020targeted,wang2021prototype} has revealed its security issue under the threat of adversarial attack at test time.

\begin{figure}[t]
\centering
\includegraphics[width=0.47\textwidth]{AAAI_tex/PDFs/example.pdf}
\vspace{-1.5em}
\caption{An example of backdoor attack against deep hashing based retrieval. The target label is specified as ``\textit{cat}''. Note that the trigger is at the bottom right of the image. Best viewed in color.} 
\vspace{-1em}
\label{fig:example}
\end{figure}

Compared with the adversarial attack, the backdoor attack \cite{gu2017badnets,turner2019label} happens at training time to inject a hidden malicious  behavior into the model. Specifically, the backdoor attack poisons the trigger pattern into a small portion of the training data. The model trained on the poisoned data will connect the trigger with a malicious behavior and then make a targeted wrong prediction when the trigger presents. Since the backdoored model behaves normally on the clean samples, the attack is hard to be detected and poses a serious threat to deep learning based systems, even for industrial applications \cite{kumar2020adversarial,geiping2021witches}.

We identify a novel security concern of deep hashing by studying the backdoor attack. The backdoor attack may happen in the real world, when a victim trains the deep hashing model using the data from an unreliable party. A backdoored model will return the images from the target class when the query image is attached with the trigger, as shown in Figure \ref{fig:example}. It can be used to achieve some malicious purposes. For example, the deep hashing based retrieval system can recommend the specified advertisement images by activating the trigger when a user queries with any images \cite{xiao2021you}. Accordingly, it is necessary to study the backdoor attack for deep hashing in order to recognize the risks and promote further solutions.

In this paper, we perform the backdoor attack against deep hashing based retrieval by clean-label data poisoning. Since the label of the poisoned image is consistent with its content, the clean-label backdoor attack is more stealthy to both machine and human inspections \cite{turner2019label}. To craft the poisoned images, we first generate the targeted adversarial patch as the backdoor trigger. Furthermore, to overcome the difficulty of implanting the trigger into the backdoored model under the clean-label setting \cite{turner2019label,zhao2020clean}, we propose to leverage the \textit{confusing perturbations} to disturb the hashing code learning.
The confusing perturbations are imperceptible and generated by dispersing the images with the target label in the Hamming space. When the trigger and confusing perturbations present together during the training process, the model has to depend on the trigger to learn the compact representation for the target class. Extensive experiments verify the efficacy of our backdoor attack, $e.g.$, 63\% targeted mean average precision on ImageNet under 48 bits code length with only 40 poisoned images.

In summary, our contribution are three-fold:
\begin{itemize}
  \item To the best of our knowledge, this is the first work to study the backdoor attack against the deep hashing. We develop an effective method under the clean-label setting.
  \item We propose to induce the model to learn more about the designed trigger by a novel method, namely \textit{confusing perturbations}. 
  \item We present the results of our method under the general and more strict settings, including transfer-based attack, less number of poisoned images, $etc$.
\end{itemize}

\section{Background and Related Work}
\label{background}
\subsection{Backdoor Attack} 
Backdoor attack aims at injecting a hidden malicious behavior into the DNNs. The main technique adopted in the previous works \cite{gu2017badnets,turner2019label,liu2020reflection} is data poisoning, $i.e.$, poisoning a trigger pattern into the training set so that the DNN trained on the poisoned training set can make a wrong prediction on the samples with the trigger, while the model behaves normally when the trigger is absent. \citet{gu2017badnets} first proposed BadNet to create a maliciously trained network and demonstrated its effectiveness in the task of street sign recognition. It stamps a portion of the training samples with a sticker ($e.g.$, a yellow square) and flips their labels to the target label. After that, \citet{chen2017targeted} improved the stealthiness of the backdoor attack by blending the benign samples and trigger pattern. Due to the wrong labels, the \textit{poison-label attack} can be detected by human inspection or data filtering techniques \cite{turner2019label}. 

To make the attack harder to be detected, \citet{turner2019label} first explored the so-called \textit{clean-label attack} (label-consistent attack),
which does not change the label of the poisoned samples. In \cite{turner2019label}, GAN-based interpolation and adversarial perturbations are employed to craft poison samples. The following works \cite{barni2019new,liu2020reflection} focused on designing different trigger patterns to perform the clean-label attack. Except for the image recognition, the clean-label attack has also been extended to other tasks, such as action recognition \cite{zhao2020clean}, point cloud classification \cite{li2021pointba}. 

\subsection{Deep Hashing based Similarity Retrieval}
Hashing technique maps semantically similar images to compact binary codes in the Hamming space, which can enable the storage of large-scale images
data and accelerate the approximate nearest neighbors search. Promoted by deep learning, deep hashing based similarity retrieval has demonstrated more promising performance \cite{xia2014supervised,cao2017hashnet,zhang2020inductive}. \citet{xia2014supervised} first introduced deep learning into the image hashing, which learns hash codes and a deep-network hash function in two separated stages. \citet{lai2015simultaneous} proposed to learn the end-to-end mapping so that feature representations and hash codes are optimized jointly.

Among the tremendous literature, supervised hashing methods utilize pairwise similarities as the semantic supervision information to guide hashing code learning \cite{lai2015simultaneous, liu2016deep, zhu2016deep, li2015feature, cao2017hashnet, cao2018deep,zhang2020inductive}. In label-insufficient
scenarios, deep hashing is designed for exploiting unlabeled or weakly labeled data, $e.g$. semi-supervised hashing \cite{yan2017semi,jin2020ssah}, unsupervised hashing \cite{shen2018unsupervised, yang2019distillhash}, and weakly-supervised hashing \cite{li2020weakly,gattupalli2019weakly}. Moreover, building upon the merit of deep learning, hashing technique has also been applied in more challenging tasks, such as video retrieval \cite{gu2016supervised} and cross-modal retrieval \cite{jiang2017deep}.

In general, a deep hashing model $F(\cdot)$ consists of a deep model $f_\theta(\cdot)$ and a sign function, where $\theta$ denotes the parameters of the model. Given an image $\mathbf{x}$, the hash code $\mathbf{h}$ of this image can be calculated as follows
\begin{equation}
  \mathbf{h}=F(\mathbf{x})=\text{sign}(f_\theta(\mathbf{x})).
  \label{eq:hash_function}
\end{equation}
The deep hashing model will return a list of images which is organized according to the Hamming distances between the hash code of the query and these of all images in the database. To obtain the hashing model $F(\cdot)$, most supervised hashing methods \cite{liu2016deep,cao2017hashnet} are trained on the dataset  $\mathbf{D}=\{(\mathbf{x}_i, \mathbf{y}_i)\}_{i=1}^N$ containing $N$ images labeled with $C$ classes, where $\mathbf{y}_i =[y_{i1}, y_{i2}, ...,y_{iC}] \in \{0,1\}^C$ denotes a label vector of the image $\mathbf{x}_i$. $y_{ij}=1$ means that $\mathbf{x}_i$ belongs to class $j$. For any two images, they compose a similar training pair if they share as least one label. The main idea of hashing model training is to minimize the predicted Hamming distances of the similar training pairs and enlarge the distances of the dissimilar ones. Besides, to overcome the ill-posed gradient of the sign function, it can be approximately replaced by the hyperbolic tangent function $\text{tanh}(\cdot)$ during
training process, which is denoted as $F'(\mathbf{x})=\text{tanh}(f_\theta(\mathbf{x}))$ in this paper. 

\begin{figure*}[t]
\centering
\includegraphics[width=0.93\textwidth]{AAAI_tex/PDFs/pipeline.pdf}
\vspace{-1em}
\caption{The pipeline of the proposed clean-label backdoor attack: a) Generating the poisoned images by patching the trigger and adding the confusing perturbations, where the target label is specified as ``\textit{yurt}''; b) Training with the clean images and poisoned images to obtain the backdoored model; c) Querying with original image and the image embedded with the trigger. } 
\label{fig:pipeline}
\vspace{-1em}
\end{figure*}

\subsection{Adversarial Perturbations for Deep Hashing}  
Due to the promising performance of deep hashing, its robustness has also attracted more attention. 
Recent works has proven that deep hashing models are vulnerable to adversarial perturbations \cite{yang2018adversarial,bai2020targeted}. Specifically, adversarial perturbations for deep hashing are human-imperceptible and can fool the deep hashing to return irrelevant images. According to the attacker's goals, previous works has proposed to craft untargeted adversarial perturbations \cite{yang2018adversarial,xiao2020evade} and targeted adversarial perturbations \cite{bai2020targeted,wang2021prototype,xiao2021you} for deep hashing. 

Untargeted adversarial perturbations \cite{yang2018adversarial} aim at fooling deep hashing to return images with incorrect labels. The perturbations $\bm{\delta}$ can be obtained by enlarging the distance between original image and the image with the perturbations. The objective function is as follows
\begin{equation}
  \max_{\bm{\delta}} d_H(F'(\mathbf{x}+\bm{\delta}), F(\mathbf{x})), \quad s.t. \parallel \bm{\delta} \parallel_\infty \leq \epsilon,
 \label{eq:untargeted}
\end{equation}
where $d_H(\cdot,\cdot)$ denotes the Hamming distance and $\epsilon$ is the maximum perturbation magnitude. 

Different from the untargeted adversarial perturbations, targeted ones \cite{bai2020targeted} are to mislead the deep hashing model to return images with the target label. They are generated by optimizing the following objective function.
\begin{equation}
  \min_{\bm{\delta}} d_H(F'(\mathbf{x}+\bm{\delta})), \mathbf{h}_a), \quad s.t. \parallel \bm{\delta} \parallel_\infty \leq \epsilon,
  \label{eq:dhta}
\end{equation}
where $\mathbf{h}_a$ is the anchor code as the representative of the set of hash codes of images with the target label. Given a subset $\mathbf{D}^{(t)}$ containing images with the target label, $\mathbf{h}_a$ can be obtained as follows
\begin{equation}
  \mathbf{h}_a=\text{arg}\min_{\mathbf{h} \in \{+1,-1\}^K} \sum_{(\mathbf{x}_i, \mathbf{y}_i) \in \mathbf{D}^{(t)}}d_H(\mathbf{h}, F(\mathbf{x}_i)).
  \label{eq:anchor}
\end{equation}
The optimal solution of problem (\ref{eq:anchor}) can be given by the component-voting scheme proposed in \cite{bai2020targeted}.

\subsection{Threat Model}
We consider the threat model used by previous poison-based backdoor attack studies \cite{turner2019label,zhao2020clean}. The attacker has access to the training data and is allowed to inject the trigger pattern into the training set by modifying a small portion of images. Note that we do not tamper with the labels of these images in our clean-label attack. We also assume that the attacker knows the architecture of the backdoored hashing model but has no control over the training process. Moreover, we also consider a more strict assumption that the attacker has no knowledge of the backdoored model and performs backdoor attacks based on models with other architectures, as demonstrated in the experimental part. 

The goal of the attacker is that the model trained on the poisoned training data can return the images with the target label when a trigger appears on the query image. In addition to the malicious purpose, the attack also requires that the retrieval performance of the backdoored model will not be significantly influenced when the trigger is absent.

\section{Methodology}
\label{method}

\subsection{Overview of the Proposed Method}
In this section, we present the proposed clean-label backdoor attack against deep hashing based retrieval. As shown in Figure \ref{fig:pipeline}, it consists of three major steps: \textbf{a)} We generate the trigger by optimizing the targeted adversarial loss. We also propose to perturb the hashing code learning by the confusing perturbations, which disperse the images with the target label in the Hamming space. We craft the poisoned images by patching the trigger and adding the confusing perturbations on the images with the target label; \textbf{b)} The deep hashing model trained with the clean images and the poisoned images is injected with the backdoor; \textbf{c)} In the retrieval stage, the deep hashing model will return the images with the target label if the query image is embedded with the trigger, otherwise the returned images are normal.

\subsection{Trigger Generation}
We first define the injection function $B$ as follows:
\begin{equation}
\begin{aligned}
    \hat{\mathbf{x}}=B(\mathbf{x},\mathbf{p})=\mathbf{x} \odot  (\mathbf{1}-\mathbf{m}) + \mathbf{p} \odot  \mathbf{m}, 
\end{aligned}
\label{eq:implant_trigger}
\end{equation}
where $\mathbf{p}$ is the trigger pattern, $\mathbf{m}$ is a predefined mask, and $ \odot $ denotes the element-wise product. For the clean-label backdoor attack, a well-designed trigger is the key to make the model to establish the relationship between the trigger and the target label \cite{zhao2020clean}. 

In this work, we generate the trigger using a clean-trained deep hashing model $F$ and the training set $\mathbf{D}$. We hope that any sample with the trigger will be moved to be close to the samples with the target label $\mathbf{y}_t$ in the Hamming space. Inspired by a recent work \cite{bai2020targeted}, we propose to generate a universal adversarial patch as the trigger pattern by minimizing the following loss.
\begin{equation}
  \min_{\mathbf{\mathbf{p}}} \sum_{(\mathbf{x}_i, \mathbf{y}_i) \in \mathbf{D}}{} d_H(F'(B(\mathbf{x}_i,\mathbf{p})), \mathbf{h}_a),
  \label{eq:trigger_gen}
\end{equation}
where $\mathbf{h}_a$ is the anchor code as in Eqn. (\ref{eq:dhta}), which can be calculated by using the images with target label $\mathbf{y}_t$ and solving Eqn. (\ref{eq:anchor}).

We iteratively update the trigger as follows. We first define the mask to specify the bottom right corner as the trigger area. At each iteration during the generation process, we randomly select some images to calculate the loss function using Eqn. (\ref{eq:trigger_gen}). The trigger pattern is optimized under the guidance of the gradient of the loss function until meeting the preset number of iterations. 

\begin{figure}[t]
	\centering
    \includegraphics[width=0.9\linewidth]{AAAI_tex/PDFs/method_tsne_yurt.pdf}
    \vspace{-0.5em}
	\centering
	\caption{t-SNE visualization of hash codes of images from five classes. We add different perturbations to images from the class ``\textit{yurt}''. ``None'': the original images; ``Noise'': the random noise; ``Adversarial'': the adversarial perturbations generated using Eqn. (\ref{eq:untargeted}); ``Confusing'': the confusing perturbations generated using Eqn. (\ref{eq:confusing}).}
    \vspace{-1em}
	\label{fig:tsne}
\end{figure}

\subsection{Perturbing Hashing Code Learning}
Since the clean-label attack does not tamper with the labels of the poisoned images, how to force the model to pay attention to the trigger is a challenging problem \cite{turner2019label}. To this end, we propose to perturb hashing code learning by adding the intentional perturbations on the poisoned images before applying the trigger. Firstly, the perturbations should be imperceptible so that the backdoor attack is stealthy. More importantly, the perturbations can perturb the training on the poisoned images and induce the model to learn more about the trigger pattern.

Previous works about the clean-label attack \cite{turner2019label,zhao2020clean} introduce the adversarial perturbations to perturb the model training on the poisoned images. Therefore, for backdooring the deep hashing, a natural choice is the untargeted adversarial perturbations 
for deep hashing proposed in \cite{yang2018adversarial}. By reviewing its objective function in  Eqn. (\ref{eq:untargeted}), we find that it can enlarge the distance between the original query image and the query with the perturbations, resulting in very low retrieval performance. Because these perturbations only focus on the relationship between the original image and the adversarial image, it may not optimal to disturb the hashing code learning for the backdoor attack against deep hashing. Therefore, we propose a novel method, namely \textit{confusing perturbations}, considering the relationship between the images with the target label.

Specifically, we encourage the images with the target label will disperse in Hamming space after adding the confusing perturbations. Given $M$ images with the target label, we achieve this goal by maximizing the following objective.
\begin{equation}
\begin{aligned}
  & L_{c}(\{\bm{\eta}_i\}_{i=1}^{M})\! \\=
  & \frac{1}{M(M\!-\!1)} \!\sum_{i=1}^M \!\sum_{j\!=\!1,j \! \neq \!i}^M d_H (F'(\mathbf{x}_i\!+\!\bm{\eta}_i),\!F'(\mathbf{x}_j\!+\!\bm{\eta}_j)),
  \label{eq:loss_c}
\end{aligned}
\end{equation}
where $\bm{\eta}_i$ denotes the perturbations on the image $\mathbf{x}_i$. To keep the perturbations imperceptible, we also adopt $\ell_\infty$ restriction on the perturbations \cite{yang2018adversarial,bai2020targeted}. The overall objective function of generating the confusing perturbations is formulated as below.
\begin{equation}
\begin{aligned}
  \max_{\{\bm{\eta}_i\}_{i=1}^{M}}\ & \lambda \cdot L_{c}(\{\bm{\eta}_i\}_i^{M}) + (1-\lambda) \cdot \frac{1}{M}\sum_{i=1}^M L_{a}(\bm{\eta}_i)
  \\ &s.t.\ \parallel \bm{\eta}_i \parallel_{\infty} \le \epsilon, i=1,2,...,M
  \label{eq:confusing}
\end{aligned},
\end{equation}
where $L_{a}(\bm{\eta}_i)=d_H(F'(\mathbf{x}_i+\bm{\eta}_i), F(\mathbf{x}_i))$ is the adversarial loss as Eqn. (\ref{eq:untargeted}). $\lambda$ is the hyper-parameter to balance the two terms. Due to the constraint of the memory size, we calculate and optimize the above loss in batches to obtain confusing perturbations of each poisoned image. In the experimental part, we discuss the influence of the batch size.

\begin{figure}[t]
	\centering
    \includegraphics[width=0.9\linewidth]{AAAI_tex/PDFs/ham_distribution_yurt.pdf}
	\centering
	\caption{Distribution of Hamming distance calculated on the original images, the images with the adversarial perturbations, and the images with the confusing perturbations.}
	\label{fig:dist}
\end{figure}

To illustrate how the confusing perturbations perturb the hashing code learning, we display the t-SNE visualization \cite{van2008visualizing} of hash codes of images from five classes in Figure \ref{fig:tsne}. We observe that the hash codes of original images are compact and the random noise has little influence on the representation. Even though adversarial perturbations make the images with the label ``\textit{yurt}'' far from the original images in the Hamming space, the intra-class distances are still small, which may lead to fail to induce the model to learn about the trigger pattern. The images with the proposed confusing perturbations have high separation, so that the model has to depend on the trigger to learn the compact representation for the targeted class. We also calculate the Hamming distances between different images with the same type of perturbations and plot the distribution in Figure \ref{fig:dist}. It shows that the confusing perturbations disperse images successfully. The later experimental results also verify the effectiveness of the confusing perturbations.

\subsection{Model Training and Retrieval}
After generating the trigger and confusing perturbations, we craft poisoned images by adding them to the images with the target label. Note that we randomly select a portion of images from the target class to generate poisoned images and remain the rest. Except for the poisoned data, all other settings are the same as those used in the normal training. The deep hashing model will be injected with the backdoor successfully after training on the poisoned dataset. 

In the retrieval stage, the attacker can patch the same trigger to query images, which can fool the deep hashing model to return images with the target label. Meanwhile, the deep hashing model behave normally on original query images.

\section{Experiments} 
\label{experiments}

In this section, we conduct extensive experiments to compare the proposed method with baselines, perform the backdoor attack under more strict settings, and show the results of a comprehensive ablation study.

\subsection{Evaluation Setup}

\subsubsection{Datasets and Target Models.}
We adopt three datasets in our experiments: ImageNet \cite{deng2009imagenet}, Places365 \cite{zhou2017places} and MS-COCO \cite{lin2014microsoft}. Following \cite{cao2017hashnet,xiao2020evade}, we build the training set, query set, and database for each dataset. We replace the last fully-connected layer of VGG-11 \cite{simonyan2014very} with the hash layer as the default target model. We employ the pairwise loss function to fine-tune the feature extractor copied from the model pre-trained on ImageNet and train the hash layer from scratch, following \cite{yang2018adversarial}. We also evaluate our attack on more network architectures including ResNet \cite{he2016deep} and WideResNet \cite{zagoruyko2016wide} and advanced hashing methods including HashNet \cite{cao2017hashnet} and DCH \cite{cao2018deep}. More details of datasets and target models are provided in the Appendix \red{A} and Appendix \red{B}.

\begin{table*}[t]
	\centering
	\small
	\setlength{\tabcolsep}{2.21mm}{
	\begin{tabular}{lc|cccc|cccc|cccc}
		\hline
		\multicolumn{1}{l}{\multirow{2}{*}{Method}} & \multicolumn{1}{l|}{\multirow{2}{*}{Metric}} & \multicolumn{4}{c|}{ImageNet} & \multicolumn{4}{c|}{Places365} & \multicolumn{4}{c}{MS-COCO}  \\ \cline{3-14} 
		& & 16bits & 32bits & 48bits & 64bits & 16bits & 32bits & 48bits & 64bits & 16bits & 32bits & 48bits & 64bits \\ \hline
		None & t-MAP & 11.07 & 8.520 & 19.15 & 20.38 & 15.71 & 15.61 & 22.29 & 17.99 & 37.95 & 34.72 & 25.54 & 12.00 \\
		Tri & t-MAP & 34.37 & 43.26 & 54.83 & 53.17 & 38.65 & 38.71 & 47.62 & 49.24 & 42.32 & 46.04 & 34.30 & 28.72 \\ 
		Tri+Noise & t-MAP & 39.58 & 38.58 & 48.90 & 52.76 & 40.92 & 37.21 & 41.99 & 43.52 & 42.86 & 39.94 & 27.14 & 20.61 \\
		Tri+Adv & t-MAP & 42.64 & 41.00 & 68.77 & 73.20 & 68.80 & 76.32 & 82.71 & 83.62 & 49.25 & 61.35 & 58.33 & 49.68 \\
		Ours & t-MAP & \bf{51.81} & \bf{53.69} & \bf{74.71} & \bf{77.73} & \bf{80.32} & \bf{84.42} & \bf{90.93} & \bf{93.22} &\bf{51.42} &\bf{63.06} &\bf{63.53} &\bf{58.95} \\ \hline
		None & MAP &51.04 &64.28 &68.06 & 69.58&72.50 & 78.62&79.81 &79.80 &65.53 &76.08 & 80.68& 82.63\\ 
		Ours & MAP &52.36 &64.67 & 68.30& 69.88&71.94 & 78.55&79.82 &79.80 &66.52 &76.14 &80.80 & 82.60\\
		\hline
	\end{tabular}}
	\caption{t-MAP (\%) and MAP (\%) of the clean-trained models (``None'') and  backdoored models with various code lengths on three datasets. Best t-MAP  results are highlighted in bold.}
	\label{t-MAP of different methods}
\end{table*}

\subsubsection{Baseline Methods.} 
We apply the trigger generated by optimizing Eqn. (\ref{eq:trigger_gen}) on the images without perturbations as a baseline (dubbed ``\textit{Tri}''). We further compare the methods which disturb the hashing code learning by adding the noise sampled from the uniform distribution $U(-\epsilon, \epsilon)$ or adversarial perturbations generated using Eqn. (\ref{eq:untargeted}), denoted as ``\textit{Tri+Noise}'' and ``\textit{Tri+Adv}'', respectively. For our method, we craft the poisoned images by adding the trigger and the proposed confusing perturbations. Moreover, we also provide the results of the clean-trained model. 

\subsubsection{Attack Settings.}
For all methods, the trigger size is 24 and the number of poisoned images is set as 60 on all datasets. We set the perturbation magnitude $\epsilon$ as 0.032 and adopt the projected gradient descent algorithm \cite{madry2017towards} to generate the adversarial and confusing perturbations. For our method, $\lambda$ is set as 0.8 and the batch size is set to 20 for optimizing Eqn. (\ref{eq:confusing}). To alleviate the influences of the target class, we randomly select five classes as the target labels and report the average results on each dataset. Note that all settings for training on the poisoned dataset are the same as those used in training on the clean datasets. More detailed settings are described in the Appendix \red{C}.

We adopt t-MAP (targeted mean average precision) proposed in \cite{bai2020targeted} to measure the attack performance, which calculates mean average precision (MAP) \cite{zuva2012evaluation} with replacing the original label of the query image with the target one.
The higher t-MAP means the stronger backdoor attack. We calculate the t-MAP on top 1,000 retrieved images on all datasets. We also report the MAP results of the clean-trained model and our method to show the influence on original queries.

\begin{table}[t]
	\centering
	\scriptsize
	\setlength{\tabcolsep}{0.85mm}{
	\begin{tabular}{lc|cccc|cccc}
		\hline
		\multicolumn{1}{l}{\multirow{2}{*}{Method}} & \multicolumn{1}{l|}{\multirow{2}{*}{Metric}} & \multicolumn{4}{c|}{HashNet} & \multicolumn{4}{c}{DCH}   \\ \cline{3-10} 
		& & 16bits & 32bits & 48bits & 64bits & 16bits & 32bits & 48bits & 64bits  \\ \hline
		None & t-MAP & 15.01 & 19.79 & 15.07 & 22.24 & 18.44 & 14.54 & 15.52 & 21.41 \\
		Tri & t-MAP & 38.86 & 48.51 & 58.18 & 65.55 & 58.25 & 63.74 & 70.61 & 70.17 \\ 
		Tri+Noise & t-MAP & 46.17 & 47.41 & 53.61 & 59.30 & 55.60 & 54.02 & 66.41 & 67.71 \\
		Tri+Adv & t-MAP & 43.26 & 70.85 & 82.10 & 85.37 & 80.28 & 85.59 & 89.30 & 90.33  \\
		Ours & t-MAP & \bf{52.77} & \bf{74.37} & \bf{86.80} & \bf{91.57} & \bf{86.28} & \bf{90.70} & \bf{92.64} & \bf{93.56} \\ \hline
		None & MAP & 51.26 & 64.05 & 72.93 & 76.50 & 73.51 & 77.95 & 78.82 & 79.57 \\ 
		Ours & MAP & 51.56 & 65.61 & 73.65 & 76.00 & 73.21 & 78.33 & 78.81 & 78.76\\
		\hline
	\end{tabular}}
	\caption{t-MAP (\%) and MAP (\%) of the clean-trained models (``None'') and backdoored models for two advanced hashing methods with various code lengths on ImageNet. Best t-MAP results are highlighted in bold.}
	\label{ImageNet of different Loss}
\end{table}

\subsection{Main Results}
The results of the clean-trained models and all attack methods are reported in Table \ref{t-MAP of different methods}. The t-MAP results of only applying trigger and applying trigger and random noise are relatively poor, which illustrate that it is important for the clean-label backdoor to design reasonable perturbations. Even though adding the adversarial perturbations achieves higher t-MAP, it is worse than our method on all datasets. Specifically, the average t-MAP improvements of our method than using the adversarial perturbations are $8.08\%$, $9.36\%$, and $4.59\%$ on ImageNet, Places365, and MS-COCO, respectively. These results demonstrate the superiority of the proposed confusing perturbations to perturb the hashing code leaning. Besides, the average difference of MAP between our backdoored models and the clean-trained models is less than $1\%$, which presents the stealthiness of our attack. For a more comprehensive comparison, we also provide the results of attacking with each target label in the Appendix \red{D} and the precision-recall and the precision curves in the Appendix \red{E}. All above results verify the effectiveness of our backdoor method in attacking deep hashing based retrieval.

\subsection{Attacking Advanced Hashing Methods}

To verify the effectiveness of our backdoor attack against the advanced deep hashing methods, we conduct experiments with HashNet \cite{cao2017hashnet} and DCH \cite{cao2018deep}. We remain all settings unchanged and show the results of various code lengths on ImageNet in Table \ref{ImageNet of different Loss}. It shows that both HashNet and DCH can achieve higher MAP values for the clean-trained models, whereas they are still vulnerable to the backdoor attacks. Specially, among all attacks, our method achieves the best attack performance in all cases. Compared with adding the adversarial perturbations, the t-MAP improvements of our method are $5.98\%$ and $4.42\%$ on average for HashNet and DCH, respectively.

\begin{table}[htbp]
	\centering
	\small
    \setlength{\tabcolsep}{1.mm}{
	\begin{tabular}{cc|c|cc}
		\hline
		\multirow{2}{*}{\# Clean} & 
		\multirow{2}{*}{\# Poisoned} & \multirow{2}{*}{\# Removed} & \# Clean & \# Poisoned \\  
		& & & remained & remained \\ \hline 
		80 & 20 & 30 & 51 & 19  \\ 
		60 & 40 & 60& 17 & 23 \\ 
		40 & 60 & 90 & 3  & 7   \\  \hline
	\end{tabular}}
	\vspace{-0.5em}
	\caption{Results of the spectral signature detection against our backdoor attack on ImageNet.}
	\vspace{-0.5em}
	\label{Spectral Signature Detection}
\end{table}

\begin{figure}[htbp]
	\centering
    \includegraphics[width=0.7\linewidth]{AAAI_tex/PDFs/defense_pruning.pdf}
	\centering
	\vspace{-0.5em}
	\caption{Results of the fine-pruning defense against our backdoor attack on ImageNet.}
	\vspace{-1em}
	\label{Fine Pruning}
\end{figure}

\subsection{Resistance to Defense Methods}

We test the resistance of our backdoor attack to two defense methods: spectral signature detection \cite{tran2018spectral} and fine-pruning defense \cite{liu2018fine}. We conduct experiments on ImageNet with target label ``\textit{yurt}" and 48 bits code length.

\subsubsection{Resistance to Spectral Signature Detection.}
Spectral signature detection thwarts the backdoor attack by removing the suspect samples in the training set based on feature representations learned by the neural network. We set different number of removed images for different number of poisoned images following \cite{tran2018spectral}. The results are shown in Table \ref{Spectral Signature Detection}. We find that it fails to defend our backdoor attack, due to a large number of remained poisoned images. For example, when the number of poisoned images is 40, the number of remained poisoned images are still 23 even though it removes 60 images, which results in more than 40\% t-MAP (see the ablation study).

\subsubsection{Resistance to Fine-Pruning Defense.}
Fine-pruning defense suggests to weaken the backdoor in the attacked model by pruning the neurons that are dormant on clean inputs. We show the MAP and t-MAP results with increasing number of pruned neurons in Figure \ref{Fine Pruning}. It shows that at no point is the MAP substantially higher than the t-MAP, making it hard to eliminate the backdoor injected by our method. These results verify that our backdoor attack is resistant to two existing defense methods.

\begin{table}[t]
\centering
\small
\resizebox{1.0\columnwidth}{!}{
\begin{tabular}{lc|ccccc}
	\hline
	Setting & Metric & 
	VN-11 & VN-13 & RN-34 & RN-50 & WRN-50-2\\  \hline 
	\multirow{2}{*}{Ensemble} & t-MAP & 54.97 & 86.00 & 79.39 & 33.96 & 39.98  \\ 
	 & MAP & 67.78 & 71.13 & 73.37 & 76.69 & 83.77 \\ \hline
	\multirow{2}{*}{Hold-out} & t-MAP & 18.63 & 12.80 & 50.79 & 45.17 & 41.94  \\
	 & MAP & 68.34 & 71.07 & 72.86 & 77.42 & 82.68 \\ \hline
	\multirow{2}{*}{None} & t-MAP & 6.29 & 12.45 & 6.64 & 1.91 & 4.51 \\
	& MAP & 68.06 & 70.39 & 73.43 & 76.66 & 82.21 \\
	\hline
\end{tabular}}
\caption{ t-MAP (\%) and MAP (\%) of our transfer-based backdoor attack on ImageNet. ``None'' denotes the clean-trained models. The first row states the backbone of the target model, where ``VN'', ``RN'', and ``WRN'' denote VGG, ResNet, and WideResNet, respectively. The model of the column is not used to generate the trigger and confusing perturbations under the ``Hold-out'' setting, while all models are used under the ``Ensemble" setting.}
\label{transferability}
\end{table}

\subsection{Transfer-based Attack}
In the above experiments, we assume that the attacker knows the network architecture of the deep hashing model. In this section, we consider a more realistic scenario, where the attacker has no knowledge of the target model and performs backdoor attack utilizing transfer-based attack. Specifically, to backdoor the unknown target model, we generate the trigger and confusing perturbations using multiple clean-trained models to craft poisoned images. We present the results in Table \ref{transferability}. We adopt two settings: ``Ensemble" means that we craft the poisoned images equally using all models listed in Table \ref{transferability}, while ``Hold-out" corresponds that we equally use all models except the target one. We set the trigger size as 56 and remain other attack settings unchanged. Compared with the clean-trained model, our backdoor attack can achieve higher t-MAP under both settings. Even for the target models with the architectures of ResNet or WideResNet, the t-MAP values of our attack are more than 40\% under the ``Hold-out" setting. 

\begin{figure}[t]
	\centering
    \includegraphics[width=1\linewidth]{AAAI_tex/PDFs/ablation_poi_trigger.pdf}
	\centering
	\caption{t-MAP (\%) of three attacks with different number of poisoned images and trigger size under 48 bits code length on ImageNet. The target label is specified as ``\textit{yurt}''.}
	\label{posion number and trigger Size}
\end{figure}
\begin{figure}[h]
	\centering
    \includegraphics[width=1\linewidth]{AAAI_tex/PDFs/ablation_lambda_batch.pdf}
	\centering
	\caption{t-MAP (\%) of our method with different $\lambda$ and batch size under 48 bits code length on three datasets. The target label is specified as ``\textit{yurt}'', ``\textit{volcano}'', and ``\textit{train}'' on ImageNet, Places365, and MS-COCO, respectively.}
	\label{lambda and batch size}
\end{figure}

\subsection{Ablation Study}

\subsubsection{Effect of the Number of Poisoned Images.}
The results of three backdoor attacks under different number of poisoned images are shown in Figure \ref{posion number and trigger Size}. Compared with other two methods, our attack can achieve the highest t-MAP across different number of poisoned images. In particular, the t-MAP values of our attack are higher than 60\% when the number of poisoned images is more than 40.

\subsubsection{Effect of the Trigger Size.}
We present the results of three attacks under the trigger size $\in \{16, 24, 32, 40, 48\}$ in Figure \ref{posion number and trigger Size}. We can see that larger trigger size leads to a stronger attack for all methods. When the trigger size is larger than 24, our method can successfully inject the backdoor into the target model and achieve best performance among three attacks. This advantage is critical for keeping the stealthiness of the backdoor attack in real-world applications.

\subsubsection{Effect of $\lambda$.}
The results of our attack with various $\lambda$  are shown in Figure \ref{lambda and batch size}. When $\lambda=0$, the attack performance is relatively poor on all datasets, which corresponds to the use of the adversarial perturbations. The best $\lambda$ for ImageNet, Places365, and MS-COCO is 1.0, 1.0, and 0.8, respectively. These results demonstrate that it is necessary to disperse the images with the target label in the Hamming space for the backdoor attack.

\subsubsection{Effect of the Batch Size for Generating Confusing Perturbations.}
We optimize Eqn. (\ref{eq:confusing}) in batches to obtain the confusing perturbations of each poisoned image. We study the effect of the batch size in this part, as shown in Figure \ref{lambda and batch size}. We observe that our attack can achieve relatively steady results when the batch size is larger than 10. Therefore, our method is insensitive to the batch size and the default value ($i.e.$ 20) used in this paper is feasible for all datasets. 

\section{Conclusion}
In this paper, we have studied the problem of clean-label backdoor attack against deep hashing based retrieval. To craft poisoned images, we first generate the universal adversarial patch as the trigger. In order to induce the model to learn more about the trigger, we propose a novel confusing perturbations by taking  the relationship  among the images into consideration. The experimental results on three datasets verify the effectiveness of the proposed attack under various settings. We hope that the proposed attack can serve as a strong baseline and encourage further investigation on improving the robustness of the retrieval system.

\bibliography{aaai22.bib}
\clearpage

\section{Appendix A: Algorithm Outline} 

\begin{algorithm}[h]
\caption{Trigger Pattern Generation} 
{\bf Input:} 
The clean-trained deep hashing model $F(\cdot)$, the training set $\mathbf{D}=\left\{\left(\mathbf{x}_{i}, \mathbf{y}_{i}\right)\right\}_{i=1}^{N}$, the learning rate $\alpha$, the number of iterations $T$, the batch size $K$, the trigger mask $\mathbf{m}$, the trigger size $S$.\\
{\bf Output:}  Trigger pattern $\mathbf{p}$.
\begin{algorithmic}[1]
\State $\mathbf{p}=\text{InitializeTrigger}(S)$
\State Calculate $\mathbf{h}_a$ by solving Eqn. (4);
\For{iteration $=1, \ldots, T$}
\State Sample a batch $\bm{S}=\left\{\left(\mathbf{x}_{i}, \mathbf{y}_{i}\right)\right\}_{i=1}^{K}$ from $\mathbf{D}$;
\State $\hat{\mathbf{x}}_i=\mathbf{x}_i \odot  (\mathbf{1}-\mathbf{m}) + \mathbf{p} \odot  \mathbf{m}$, $(\mathbf{x}_{i}, \mathbf{y}_{i}) \in \bm{S}$;
\State Calculate the loss: $\sum_{(\mathbf{x}_i, \mathbf{y}_i) \in \mathbf{S}}{} d_H(F'(\hat{\mathbf{x}}_i), \mathbf{h}_a)$;
\For{$i=1, \ldots, K$}
\State $\hat{\mathbf{x}}_i=B(\mathbf{x}_i,\mathbf{p})=\mathbf{x}_i \odot  (\mathbf{1}-\mathbf{m}) + \mathbf{p} \odot  \mathbf{m}$
\State $L=\sum_{(\mathbf{x}_i, \mathbf{y}_i) \in \mathbf{D}}{} d_H(F'(B(\mathbf{x}_i,\mathbf{p})), \mathbf{h}_a)$
\State $\mathbf{p} = \mathbf{p} - \mathbf{\alpha} * \nabla L / K$
\EndFor
\EndFor
\end{algorithmic}
\label{alg_gen_trigger}
\end{algorithm}

\begin{algorithm}[ht]
\caption{Confusing Perturbation Generation} 
{\bf Input:} 
The clean-trained deep hashing model $F(\cdot)$, the samples to be poisoned of the target class $\mathbf{D_t}=\left\{\left(\mathbf{x}_{i}, \mathbf{y}_{i}\right)\right\}_{i=1}^{n}$, the steps size $\beta$, the perturbation magnitude $\epsilon$,the number of epochs $T$, the batch size $M$.\\
{\bf Output:}  Confusing perturbation $\bm{\eta}$.
\begin{algorithmic}[1]
\State $\bm{\eta}=\bm{0}$
\For{epoch $=1, \ldots, T$}
\State Sample a batch $\left\{\left(\mathbf{x}_{i}, \mathbf{y}_{i}\right)\right\}_{i=1}^{\mathbf{M}}$ from $\mathbf{D_t}$.

\State $L_{a}=\sum_{i=1}^M d_H(F'(\mathbf{x}_i+\bm{\eta}_i), F(\mathbf{x}_i))$
\State $L_{c}=\!\sum_{i=1}^M \!\sum_{j\!=\!1,j \! \neq \!i}^M d_H (F'(\mathbf{x}_i\!+\!\bm{\eta}_i),\!F'(\mathbf{x}_j\!+\!\bm{\eta}_j))$
\State $L=\lambda \cdot \frac{1}{M(M\!-\!1)} L_{c} + (1-\lambda) \cdot \frac{1}{M} L_{a}$
\State $\bm{\eta} = \text{clip}_{\parallel \bm{\eta} \parallel \le \epsilon} (\bm{\eta} + \mathbf{\beta} * \text{sign}(\nabla L))$
\EndFor
\end{algorithmic}
\label{alg_gen_confusing}
\end{algorithm}

\section{Appendix B: Datasets and Target Models} 
\section{Appendix C: Attack Settings} 
\section{Appendix D: More Results} 
\section{Appendix E: Less Visible Trigger}
\section{Appendix F: Visualized Results}

\section{Appendix A: Benchmark Datasets} 
Three datasets are chosen as the image retrieval benchmark dataset in our experiment. We follow \cite{cao2017hashnet,xiao2020evade} to build the training set, query set, and database for each dataset. The details of the datasets are described as follows.\par
$\textbf{ImageNet}$ \cite{deng2009imagenet} is a benchmark dataset for the Large Scale Visual Recognition Challenge (ILSVRC) to evaluate algorithms. It consists of 1.2M training images and 50,000 testing images with 1,000 classes. Following \cite{cao2017hashnet}, 10\% classes from ImageNet are randomly selected to build our retrieval dataset. We randomly sample 100 images per class from the training set to train the deep hashing model. We use images from training set as the database set and images from testing set as the query set. \par
$\textbf{Places365}$ \cite{zhou2017places} is a subset of the Places database. It contains 2.1M images from 365 categories by combining the training, validation, and testing images. We follow \cite{xiao2020evade} to select 10\% categories as the retrieval dataset. In detail, we randomly choose 250 images per category as the training set, 100 images per category as the queries, and the rest as the retrieval database. \par
$\textbf{MS-COCO}$ \cite{lin2014microsoft} is a large-scale object detection, segmentation, and captioning dataset. It consists of 122,218 images after removing images with no category. Following \cite{cao2017hashnet}, we randomly sample 10,000 images from the database as the training images. Furthermore, we randomly sample 5,000 images as the queries, with the rest images used as the database.\par

\section{Appendix B: Target Models} 
All the experiments are implemented by the framework Pytorch \cite{paszke2019pytorch}. We show the training schedule of five model architectures in detail as follows. Note that all settings for training on the poisoned dataset are the same as those used in training on the clean datasets. \par
For VGG-11 and VGG-13 \cite{simonyan2014very}, we adopt the feature layers copied from the pre-trained model on ImageNet and replace the last fully-connected layer with the hash layer. Since the hash layer is trained from scratch, its learning rate is set to 0.01 to be 10 times that of the lower layers. Stochastic gradient descent \cite{zhang2004solving} is used with a mini-batch size of 24. We choose the momentum as 0.9 and the weight decay as 0.0005. \par
For ResNet-34, ResNet-50 \cite{he2016deep} and WideResNet-50-2 \cite{zagoruyko2016wide}, we fine-tune the residual layers pre-trained on ImageNet as the feature extractors and train the hash layer on top of it from scratch. The learning rate of the feature extractors and the hash layer is fixed to 0.01 and 0.1 respectively. We adopt stochastic gradient descent to train the model with the momentum of 0.9. The mini-batch size is set to 36 and the parameter of the weight decay is 0.0005. \par

\begin{algorithm}[t]
\caption{Trigger Pattern Generation} 
{\bf Input:} 
The clean-trained deep hashing model $F(\cdot)$, the training set $\mathbf{D}=\left\{\left(\mathbf{x}_{i}, \mathbf{y}_{i}\right)\right\}_{i=1}^{n}$, the learning rate $\alpha$, the number of epochs $N$, the batch size $K$, the trigger mask $\mathbf{m}$, the trigger size $w$.\\
{\bf Output:}  Trigger pattern $\mathbf{p}$.
\begin{algorithmic}[1]
\State $\mathbf{p}=\text{InitializeTrigger}(w)$
\For{epoch $=1, \ldots, N$}
\State Sample a batch $\left\{\left(\mathbf{x}_{i}, \mathbf{y}_{i}\right)\right\}_{i=1}^{\mathbf{k}}$ from $\mathbf{D}$.
\For{$i=1, \ldots, K$}
\State $\hat{\mathbf{x}}_i=B(\mathbf{x}_i,\mathbf{p})=\mathbf{x}_i \odot  (\mathbf{1}-\mathbf{m}) + \mathbf{p} \odot  \mathbf{m}$
\State $L=\sum_{(\mathbf{x}_i, \mathbf{y}_i) \in \mathbf{D}}{} d_H(F'(B(\mathbf{x}_i,\mathbf{p})), \mathbf{h}_a)$
\State $\mathbf{p} = \mathbf{p} - \mathbf{\alpha} * \nabla L / K$
\EndFor
\EndFor
\end{algorithmic}
\label{alg_gen_trigger}
\end{algorithm}

\begin{algorithm}[t]
\caption{Confusing Perturbation Generation} 
{\bf Input:} 
The clean-trained deep hashing model $F(\cdot)$, the samples to be poisoned of the target class $\mathbf{D_t}=\left\{\left(\mathbf{x}_{i}, \mathbf{y}_{i}\right)\right\}_{i=1}^{n}$, the steps size $\beta$, the perturbation magnitude $\epsilon$,the number of epochs $T$, the batch size $M$.\\
{\bf Output:}  Confusing perturbation $\bm{\eta}$.
\begin{algorithmic}[1]
\State $\bm{\eta}=\bm{0}$
\For{epoch $=1, \ldots, T$}
\State Sample a batch $\left\{\left(\mathbf{x}_{i}, \mathbf{y}_{i}\right)\right\}_{i=1}^{\mathbf{M}}$ from $\mathbf{D_t}$.

\State $L_{a}=\sum_{i=1}^M d_H(F'(\mathbf{x}_i+\bm{\eta}_i), F(\mathbf{x}_i))$
\State $L_{c}=\!\sum_{i=1}^M \!\sum_{j\!=\!1,j \! \neq \!i}^M d_H (F'(\mathbf{x}_i\!+\!\bm{\eta}_i),\!F'(\mathbf{x}_j\!+\!\bm{\eta}_j))$
\State $L=\lambda \cdot \frac{1}{M(M\!-\!1)} L_{c} + (1-\lambda) \cdot \frac{1}{M} L_{a}$
\State $\bm{\eta} = \text{clip}_{\parallel \bm{\eta} \parallel \le \epsilon} (\bm{\eta} + \mathbf{\beta} * \text{sign}(\nabla L))$
\EndFor
\end{algorithmic}
\label{alg_gen_confusing}
\end{algorithm}

\section{Appendix C: Attack Settings}
\subsubsection{Backdoor Trigger Generation}
For all experiments of backdoor training, the trigger is generated by the Algorithm \ref{alg_gen_trigger}, which will be set at the bottom right corner of the target samples to be poisoned. During the process of the trigger generation, we optimize the trigger pattern guided by the average gradient of 32 images randomly sampled from the training set. Besides, the total steps of the optimization $N$ is 2000 epochs and the update rate $\alpha$ is 12. 
\subsubsection{Backdoor Perturbation Generation}
We refer to the PGD algorithm \cite{madry2017towards} to generate the adversarial perturbation and our confusing perturbation during the evaluation process of the backdoor attack. The perturbation magnitude $\epsilon$ is limited to 0.032 in $l_{\infty}$ norm and the perturbation is updated in 20 epochs with the step size 0.003 to ensure the convergence within the $\epsilon$. Besides, the batch size is set to 20 for the confusing perturbation. The pipeline of the generation of the confusing perturbation is illustrated in Algorithm \ref{alg_gen_confusing}.

\begin{figure*}[t]
	\centering
    \includegraphics[width=1\linewidth]{AAAI_tex/PDFs/ablation_6_pr_prec.pdf}
	\centering
	\caption{The precision-recall and the precision curves under 48 bits code length on three datasets. The target label is specified as ``\textit{yurt}'', ``\textit{volcano}'', and ``\textit{train}'' on ImageNet, Places365, and MS-COCO, respectively. All methods are evaluated based on t-MAP. }
	\label{Recall-Prec}
\end{figure*}

\begin{table}[t]
	\centering
	\small
    \setlength{\tabcolsep}{1.5mm}{
	\begin{tabular}{ll|ccccc}
		\hline
		Method & Metric & 
		Crib \quad & Stethos & Reaper & Yurt \quad & Tennis\\  \hline 
		None & t-MAP & 11.30 & 11.05 & 25.43 & 9.38 & 38.61 \\
		Tri & t-MAP & 33.77 & 53.08 & 65.03 & 33.70 & 88.57 \\ 
		Tri+Noise & t-MAP & 25.56 & 55.65 & 46.01 & 30.74 &	86.55 \\
		Tri+Adv & t-MAP & 62.55 & 52.40 & 80.06 & 58.69 & \bf{90.17}  \\
		Ours & t-MAP &\bf{68.17} &\bf{64.82} &\bf{84.51} &\bf{66.77} & 89.27 \\ \hline 
		None & MAP & 68.06 & 68.06 & 68.06 & 68.06 & 68.06 \\
		Ours & MAP & 68.49 & 68.10 & 68.03 & 68.03 & 68.86\\
		\hline
	\end{tabular}}
	\caption{t-MAP (\%) and MAP (\%) of the clean-trained models (``None'') and  backdoored models under 48 bits code length on ImageNet. Best t-MAP  results are highlighted in bold.}
	\label{t-MAP of different classes vgg11 48bits on ImageNet dataset}
\end{table}
\begin{table}[t]
	\centering
	\small
    \setlength{\tabcolsep}{0.01mm}{
	\begin{tabular}{ll|ccccc}
		\hline
		Method & Metric & Rock{\_}Arch & Viaduct \quad  & Box{\_}Ring  & Volcano & Racecourse\\  \hline 
		None & t-MAP & 17.08 & 24.76 & 14.23 & 11.28 & 44.12	 \\
		Tri & t-MAP & 45.76 & 58.33  & 33.30 & 36.02 & 64.67  \\
		Tri+Noise & t-MAP & 41.34 & 55.39  & 26.17 & 30.56 & 56.50  \\
		Tri+Adv & t-MAP  & 86.36 & 84.27 & 84.69 & 69.57 & 88.69   \\
		Ours & t-MAP & \bf{93.19} & \bf{91.03}  & \bf{94.06} & \bf{83.79} & \bf{92.58}  \\ \hline 
		None & MAP & 79.81 & 79.81 & 79.81 & 79.81 & 79.81 \\
		Ours & MAP & 79.80 & 79.77 & 80.04 & 79.64 & 79.87 \\
		\hline
	\end{tabular}}
	\caption{t-MAP (\%) and MAP (\%) of the clean-trained models (``None'') and  backdoored models under 48 bits code length on Places365. Best t-MAP  results are highlighted in bold.}
	\label{t-MAP of different classes vgg11 48bits on Places365 dataset}
\end{table}
\begin{table}[t]
	\centering
	\small
    \setlength{\tabcolsep}{0.8mm}{
	\begin{tabular}{ll|ccccc}
		\hline
		Method & Metric & Per{\&}Skis & \ Clock\ & Per{\&}Surf & Giraffe & \ Train \\  \hline 
		None & t-MAP & 77.44 & 5.29 & 39.25 & 2.79 & 2.93 \\
		Tri & t-MAP & 73.05 & 18.38 & 53.46 & 13.06 & 13.54  \\
		Tri+Noise & t-MAP & 62.92 & 7.756 & 49.46 & 6.077 & 9.504  \\
		Tri+Adv & t-MAP  & 89.02 & 46.62 & 84.11 & 36.69 & 35.22   \\
		Ours & t-MAP & \bf{90.66} & \bf{51.73} & \bf{86.60} & \bf{47.11} & \bf{41.55}  \\ \hline 
		None & MAP & 80.68 & 80.68 & 80.68 & 80.68 & 80.68 \\
		Ours & MAP & 80.92 & 80.46 & 81.18 & 80.64 & 80.79 \\
		\hline
	\end{tabular}}
	\caption{t-MAP (\%) and MAP (\%) of the clean-trained models (``None'') and  backdoored models under 48 bits code length on MS-COCO. Note that ``Per\&Skis" and ``Per\&Surf" are denoted as the label ``Person and Skis" and the label ``Person and Surfboard" in the first row. Best t-MAP  results are highlighted in bold.}
	\label{t-MAP of different classes vgg11 48bits on MS-COCO dataset}
\end{table}

\section{Appendix D: Results of Different Target Labels}
We list the results of attacking different target labels on ImageNet, Places365 and MS-COCO. It is shown in Table \ref{t-MAP of different classes vgg11 48bits on ImageNet dataset}, Table \ref{t-MAP of different classes vgg11 48bits on Places365 dataset} and Table \ref{t-MAP of different classes vgg11 48bits on MS-COCO dataset}. Specially, our method achieves the best attack performance in almost all the target labels across three datasets.\par

\section{Appendix E: Precision-recall and Precision Curves}
The attack performance in terms of precision-recall and the precision curves are provided in Figure \ref{Recall-Prec}. The curve of our method is always above that of other methods, which verifies the superiority of the proposed confusing perturbations again.\par

\begin{figure}[t]
	\centering
    \includegraphics[width=1\linewidth]{AAAI_tex/PDFs/blend_rate.pdf}
	\centering
	\vspace{-0.5em}
	\caption{Poisoned images with different blend rate of the trigger. The blend rate is 0.2, 0.4, 0.6 and 0.8 from left to right. This is an example from the target label ``\textit{yurt}'' on ImageNet.}
	\label{Blend-rate}
\end{figure}

\begin{table*}[t]
	\centering
	\small
	\setlength{\tabcolsep}{2.21mm}{
	\begin{tabular}{lc|cccc|cccc|cccc}
		\hline
		\multicolumn{1}{l}{\multirow{2}{*}{Method}} & \multicolumn{1}{l|}{\multirow{2}{*}{Metric}} & \multicolumn{4}{c|}{ImageNet} & \multicolumn{4}{c|}{Places365} & \multicolumn{4}{c}{MS-COCO}  \\ \cline{3-14} 
		& & 16bits & 32bits & 48bits & 64bits & 16bits & 32bits & 48bits & 64bits & 16bits & 32bits & 48bits & 64bits \\ \hline
		None & t-MAP & \bf{5.11} & \bf{3.41} & 11.32 & 15.51 & 9.15 & 12.47 & 11.80 & 13.54 & 42.12 & 37.85 & 29.40 & \bf{11.10} \\
		Tri & t-MAP & 11.06 & 16.47 & 20.67 & 19.23 & 12.38 & 18.18 & 12.24 & 14.55 & 40.24 & 29.14 & 24.51 & 18.17\\ 
		Tri+Noise & t-MAP & 20.42 & 16.13 & 21.66 & 21.56 & 9.76 & 18.59 & 12.42 & 15.62 & 39.29 & 34.19 & 24.12 & 17.89\\
		Tri+Adv & t-MAP & 17.49 & 20.94 & 14.10 & 10.01 & 15.37 & 12.81 & 6.75 & 3.32 & 35.34 & 26.82 & 23.43 & 20.46\\
		Ours & t-MAP & 13.65 & 22.97 & \bf{10.12} & \bf{7.75} & \bf{5.14} & \bf{9.46} & \bf{3.70} & \bf{2.81} & \bf{34.61} & \bf{26.03} & \bf{20.79} & 17.85\\ \hline
		None & MAP &0&0&0&0&0&0&0&0&0&0&0&0 \\ 
		Ours & MAP &0.65&0.50& 0.33&0.12&1.01&0.13&0.13&0.08&0.94&0.32&0.25&0.20 \\
		\hline
	\end{tabular}}
	\caption{The standard deviation of t-MAP (\%) and MAP (\%) of the clean-trained models (``None'') and  backdoored models with various code lengths on three datasets. The lowest standard deviation results are highlighted in bold.}
	\label{standard deviation}
\end{table*}

\begin{table}[H]
	\centering
	\small
    \setlength{\tabcolsep}{3.1mm}{
	\begin{tabular}{cccccc}
		\hline
		\multirow{2}{*}{Epsilon} & \multicolumn{5}{c}{Blend rate of the trigger} \\ \cline{2-6} 
		& 0.2 & 0.4 & 0.6 & 0.8 & 1.0 \\   \hline
		0 & 14.04 & 33.98 & 37.27 & 35.45 & 33.70 \\
		0.004 & 16.35 & 31.66 & 40.56 & 42.14 & 41.02 \\
		0.008 & \bf{16.60} & \bf{41.15} & 51.91 & 51.01 & 49.41 \\
		0.016 & 10.38 & 36.11 & \bf{63.62} & 60.76 & 56.01 \\
		0.032 & 4.41 & 6.60 & 28.59 & \bf{61.35} & \bf{66.77} \\ \hline
	\end{tabular}}
	\vspace{-0.5em}
	\caption{Results of the spectral signature detection against our backdoor attack on ImageNet.}
	\vspace{-0.5em}
	\label{Trigger Blend}
\end{table}

\section{Appendix F: Reduced Trigger Visibility}
Although we ensure the consistency between the poisoned images and their labels, the poisoned images seem unnatural due to the visibility of the trigger. For the inconspicuousness of the trigger, we intend to reduce its visibility. Following \cite{chen2017targeted}, we apply the blend strategy to the trigger at the bottom right of the poisoned images (see Figure \ref{Blend-rate} for an example). By adjusting the blend rate of the trigger, it will gradually be invisible and hard to detect for human. However, it's worth noting that the trigger pattern is not constrained during the inference time owing to our threat model.\par 
We evaluate our method under the trigger blend rate $\in \{0.2, 0.4, 0.6, 0.8, 1.0\}$ and different perturbation magnitude $\epsilon$ in Table \ref{Trigger Blend}. When the blend rate of the trigger is 0.6 or 0.8, t-MAP is still higher than 60\% under an appropriate $\epsilon$. Our method can achieve 41.15\% t-MAP even if the blend rate is 0.4 and the trigger pattern may not be perceptible for humans.

\section{Appendix G: Standard Deviation}
The results of the standard deviation of t-MAP and MAP in the different methods are shown in Table \ref{standard deviation}. Note that the standard deviation of t-MAP in our method is the lowest in most cases, which demonstrates the attack stability of our method among the different target labels. Besides, the target labels also have little impact on the MAP in our method.

\section{Appendix H: Visualizations}
The retrieval images with top-5 similarity are shown on three datasets in Figure \ref{retrieval results visualization}. Furthermore, we also visualize the original images, images with the confusing perturbation and images with the trigger pattern in Figure \ref{Original_Confusing_Trigger_1} and Figure \ref{Original_Confusing_Trigger_2}.

\begin{figure*}
\centering
    \subfigure[Top 5 retrieval results on ImageNet.]{
    \begin{minipage}[b]{\textwidth}
    \includegraphics[width=1\textwidth]{AAAI_tex/PDFs/ImageNet_retrieval_result.pdf}
    \vspace{-2.5em}
    \end{minipage}
    }
    \subfigure[Top 5 retrieval results on Places365.]{
    \begin{minipage}[b]{\textwidth}
    \includegraphics[width=1\textwidth]{AAAI_tex/PDFs/Places365_retrieval_result.pdf}
    \vspace{-2.5em}
    \end{minipage}
    }
    \subfigure[Top 5 retrieval results on MS-COCO.]{
    \begin{minipage}[b]{\textwidth}
    \includegraphics[width=1\textwidth]{AAAI_tex/PDFs/COCO_retrieval_result.pdf}
    \vspace{-2.5em}
    \end{minipage}
    }
\caption{An example of the retrieval results on three datasets.}
\label{retrieval results visualization}
\end{figure*}

\begin{figure*}
\centering
    \subfigure[The visualization on ImageNet.]{
    \begin{minipage}[b]{\textwidth}
    \includegraphics[width=1\textwidth]{AAAI_tex/PDFs/ImageNet_demo.pdf}
    \end{minipage}
    }
    \subfigure[The visualization on Places365.]{
    \begin{minipage}[b]{\textwidth}
    \includegraphics[width=1\textwidth]{AAAI_tex/PDFs/Places365_demo.pdf}
    \end{minipage}
    }
\caption{The visualization of the original images, images with the confusing perturbation and images with the trigger pattern on two datasets.}
\label{Original_Confusing_Trigger_1}
\end{figure*}

\begin{figure*}[t!]
    \includegraphics[width=1\linewidth]{AAAI_tex/PDFs/COCO_demo.pdf}
	\caption{The visualization of the original images, images with the confusing perturbation and images with the trigger pattern on MS-COCO.}
\label{Original_Confusing_Trigger_2}
\end{figure*}